\documentclass[11pt]{article}

\usepackage[preprint]{acl}
\usepackage{times}
\usepackage{latexsym}
\usepackage{amssymb}
\usepackage{amsmath}
\usepackage[T1]{fontenc}
\usepackage[utf8]{inputenc}
\usepackage[table]{xcolor}
\usepackage{microtype}
\usepackage{inconsolata}
\usepackage{arydshln}
\usepackage{graphicx}
\usepackage[dvipsnames]{xcolor}

\usepackage{enumitem}
\usepackage{booktabs}
\usepackage{multirow}
\usepackage{booktabs,multirow}
\usepackage[table]{xcolor}
\usepackage{tabularx,array}
\usepackage{amsmath}
\usepackage{array}
\usepackage{lineno}

\usepackage{makecell}
\usepackage[most]{tcolorbox}
\usepackage{siunitx}
\definecolor{CogTitle}{RGB}{110,145,200}
\definecolor{CogFrame}{RGB}{170,195,230}
\definecolor{CogBack}{RGB}{250,252,255}

\newtcolorbox{promptbox}[2][]{%
  enhanced,
  breakable,
  colback=CogBack,
  colframe=CogFrame,
  boxrule=0.9pt,
  arc=3mm,
  left=2mm,right=2mm,top=2mm,bottom=2mm,
  colbacktitle=CogTitle,
  coltitle=white,
  fonttitle=\bfseries,
  title={#2},
  center title,
  #1
}
\usepackage[most]{tcolorbox}
\usepackage{listings}

\usepackage{fancyhdr}
\title{Unleashing Spatial Reasoning in Multimodal Large Language Models \\via Textual Representation Guided Reasoning}

\author{
  \textbf{Jiacheng Hua}$^{1,2}$ \quad
  \textbf{Yishu Yin}$^{1}$ \quad
  \textbf{Yuhang Wu}$^{1}$ \\
  \textbf{Tai Wang}$^{2}$ \quad
  \textbf{Yifei Huang}$^{3,2}$ \quad
  \textbf{Miao Liu}$^{1\dagger}$\\
  $^{1}$College of AI, Tsinghua University, Beijing, China \\
  $^{2}$Shanghai Artificial Intelligence Laboratory, Shanghai, China\\
  $^{3}$The University of Tokyo, Tokyo, Japan\\
  \texttt{hjc21@mails.tsinghua.edu.cn} \quad
  \texttt{miaoliu@mail.tsinghua.edu.cn}\\
  \url{https://trace-reasoning.github.io}
}

\begin{document}

\thispagestyle{fancy}

\maketitle

\renewcommand{\thefootnote}{}
\footnotetext{$\dagger$ Corresponding author.}

\setlist[itemize]{topsep=2pt,itemsep=1pt,parsep=0pt,leftmargin=*}

\begin{abstract}
Existing Multimodal Large Language Models (MLLMs) struggle with 3D spatial reasoning, as they fail to construct structured abstractions of the 3D environment depicted in video inputs. To bridge this gap, drawing inspiration from cognitive theories of allocentric spatial reasoning, we investigate how to enable MLLMs to model and reason over text-based spatial representations of video. Specifically, we introduce \emph{Textual Representation of Allocentric Context from Egocentric Video~(TRACE)}, a prompting method that induces MLLMs to generate text-based representations of 3D environments as intermediate reasoning traces for more accurate spatial question answering. TRACE encodes meta-context, camera trajectories, and detailed object entities to support structured spatial reasoning over egocentric videos. Extensive experiments on VSI-Bench and OST-Bench demonstrate that TRACE yields notable and consistent improvements over prior prompting strategies across a diverse range of MLLM backbones, spanning different parameter scales and training schemas. We further present ablation studies to validate our design choices, along with detailed analyses that probe the bottlenecks of 3D spatial reasoning in MLLMs.
\end{abstract}
\section{Introduction}

\begin{figure*}[!t]
    \centering
    \includegraphics[width=0.98\textwidth]{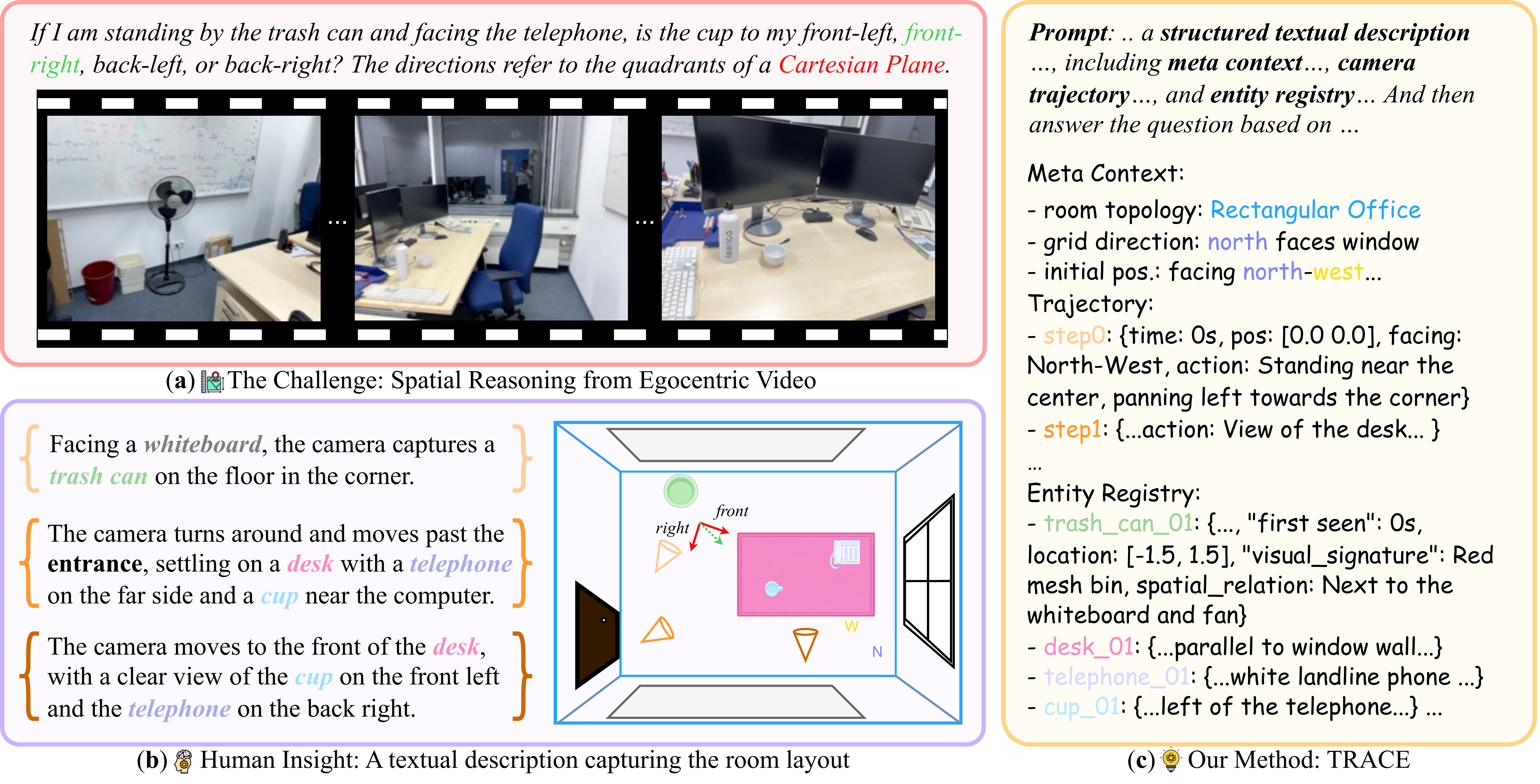}
    \caption{\textit{Motivation for Textual Representation of Allocentric Context from Egocentric Video~(TRACE) in video-based spatial reasoning.} (a) An egocentric video paired with a query that requires holistic spatial reasoning. (b) A textual description that vividly captures the room layout needed to solve the example spatial question answering (QA). (c) TRACE encodes meta-context, camera trajectory, and entities, serving as an intermediate reasoning trace for spatial QA with MLLMs.}
    \vspace{-6pt}
    \label{fig:teaser}
\end{figure*}

Cognitive science studies suggest that human reasoning about the 3D world relies on cortical mechanisms that transform visual input into hierarchical representations of objects and spatial relations, rather than operating directly on pixel-level stimuli~\cite{marr1978representation}. For instance, when humans approach the spatial reasoning question shown in Fig.~\ref{fig:teaser}(a), the solving process does not simply involve searching for cues within individual egocentric frames. Instead, we construct an immersive allocentric representation of the scene~\cite{klatzky1998allocentric}, mentally situating ourselves within the environment and reasoning about the underlying room layout to complement egocentric observations. Moreover, such allocentric representations can be vividly described using text alone, as demonstrated in Fig.~\ref{fig:teaser}(b). This observation naturally motivates the design of effective text-based video representations to enhance the spatial reasoning capabilities of existing Multimodal Large Language Models (MLLMs).

Recent studies show that existing MLLMs struggle with 3D spatial question answering (QA)~\cite{yang2025mmsi,lin2025ostbench,yang2025thinking}, despite being pretrained on massive video datasets that inherently encode rich spatial information. One key reason is that these models often overly rely on 2D visual signals and learn spurious shortcut correlations from implicit spatial cues, rather than building hierarchical abstractions of the 3D scene. In this context, we raise a fundamental scientific question: Can MLLMs be guided to explicitly construct and reason over structured allocentric representations of 3D spatial environments from 2D visual observations?

Previous work on spatially aware MLLMs generally falls into two main directions: 1) curating large-scale supervised fine-tuning data for spatial reasoning QA~\cite{daxberger2025mm,ray2024sat}, which limits scalability and generalization; or 2) incorporating additional geometric or stereo modalities into MLLMs~\cite{cheng2024spatialrgpt,zhu2024llava}, which increases system complexity and restricts applicability to off-the-shelf MLLMs. Our work explores a distinct formulation: inspired by prior approaches that extract textual descriptions from images or videos and then leverage only LLMs for VQA~\cite{wang2024videotree,fan2025videoagent}, as well as Chain-of-Thought prompting methods~\cite{wei2022chainofthought}, we propose to employ textual descriptions of 3D spatial structure as an intermediate reasoning trace that enables structured spatial reasoning in MLLMs.

Specifically, we introduce \textbf{TRACE}, short for \textbf{T}extual \textbf{R}epresentation of \textbf{A}llocentric \textbf{C}ontext from \textbf{E}gocentric Video, a prompting method that encourages MLLMs to generate a text-based allocentric representation of the 3D environment, facilitating spatial reasoning over the input egocentric video. As illustrated in Fig.~\ref{fig:teaser}(c), our proposed TRACE adopts a structured design that integrates \emph{Meta Context} describing the room layout and coordinate system, camera \emph{Trajectory} sampled over temporal windows, and explicit object \emph{Entity Registry}. This design encourages MLLMs to perform explicit reasoning over a structured allocentric representation of the scene prior to answer generation.

We conduct extensive experiments on VSI-Bench~\cite{yang2025thinking} and OST-Bench~\cite{lin2025ostbench} to evaluate TRACE, demonstrating clear performance gains over prior prompting strategies. Comparisons with other text-based video spatial representations further validate the effectiveness of our approach. We also perform detailed ablation studies and decompositional analyses to probe the bottlenecks of 3D spatial reasoning. These results highlight structured textual allocentric representations as an effective intermediate reasoning interface for video-based spatial QA in MLLMs.

\section{Related Work}

\paragraph{Spatial Representation}
Prior work has extensively studied spatial reasoning with vision–language models ~\cite{johnson2017clevr,yang2019spatialsense,hudson2019gqa}. In addition, a significant body of work has examined vision–language models in embodied or navigation-oriented settings~\cite{anderson2018vision,chen2019touchdown,shridhar2020alfred}. More recent work seeks to augment vision–language models with explicit 3D or geometric modalities~\cite{hong20233d,cheng2024spatialrgpt,zhu2024llava}, or with instruction tuning using carefully constructed data pipelines~\cite{chen2024spatialvlm,daxberger2025mm,ray2024sat}. Meanwhile, several diagnostic studies highlight that, despite these advances, current MLLMs still struggle to internally organize spatial information, motivating representations that more explicitly expose scene structure to the model~\cite{wang2024picture,liao2024reasoning}.

Our work is most closely related to recent efforts that investigate 3D spatial reasoning in MLLMs through the lens of \emph{intermediate representations for capturing scene structure}~\cite{yang2025thinking,wang2024picture}. Thinking in Space~\cite{yang2025thinking} shows that explicitly externalizing a spatial representation—such as a cognitive map—can substantially improve spatial reasoning performance, whereas standard chain-of-thought prompting alone provides limited benefit. Complementarily, SpatialEval~\cite{wang2024picture} reveals that even strong multimodal LLMs often fail to construct consistent internal 3D representations and instead rely on shortcut correlations inherited from 2D pretraining. In contrast to introducing new geometric inputs, architectural modules, or large-scale spatial instruction tuning, we propose a text-based spatial representation that serves as an intermediate reasoning step to enhance the spatial reasoning capabilities of MLLMs. Hence, our approach is flexible and broadly applicable to off-the-shelf MLLMs.

\paragraph{Text-based Description of Video}
Textual description generation for video sequences has been extensively studied. Early models addressed video captioning using sequence-to-sequence and CNN-RNN architectures~\cite{venugopalan2015sequence,donahue2015long}; later efforts focused on dense event captioning and paragraph-level video storytelling~\cite{krishna2017dense,li2018jointly,wang2021end}; another direction explored large-scale video-language pretraining for downstream tasks like retrieval and QA~\cite{sun2019videobert,luo2020univl,xu2021videoclip,lei2021less,zhao2023learning,yang2023vid2seq}. 

Our work is more closely related to approaches that build structured textual representations of video content for LLM–based question answering~\cite{wang2024videotree,wang2024videoagent,huang2025building,ren2025videorag,li2025graph,kahatapitiya2025language}. These approaches treat linguistic descriptions as the primary medium for long-context video comprehension, rather than reasoning directly over raw frames. VideoTree~\cite{wang2024videotree} builds a query-adaptive hierarchical tree of video segments and associated captions to support long-video QA with LLMs. VideoAgent~\cite{wang2024videoagent} uses an LLM as an agent to iteratively select informative clips/frames and maintain a running textual state for long-form video understanding. Video Mind Palace~\cite{huang2025building} constructs environment-grounded semantic graphs from videos as a persistent memory structure that an LLM can read for long-range reasoning. Instead of optimizing evidence coverage and retrieval over long temporal contexts, we focus on designing textual representations that enable MLLMs to explicitly reason over 3D geometry cues.

\paragraph{Prompting in M/LLM}
Prompting has become a primary inference-time mechanism for steering large M/LLM, including (i) rationale-based reasoning prompts~\cite{wei2022chainofthought,kojima2022large} (ii) decomposition and planning prompts that solve problems via sub-goals~\cite{khot2022decomposed,zhou2023leasttomost,wang2023plan,press2023measuring} (iii) aggregation and search-style prompts to reduce variance and explore alternatives~\cite{wang2023selfconsistency,yao2023treeofthought,besta2024graph} (iv) iterative self-improvement prompts via reflection~\cite{gou2023critic}. Another view is to treat language as an interface to external resources, using tool-augmented prompts and retrieval-mediated learning~\cite{yao2022react,schick2023toolformer,press2023measuring,trivedi2023interleaving}. Inspired by prior work, we propose \emph{TRACE}, the first prompting-based method that unleashes the spatial reasoning capability of MLLMs.

\section{Method}
\label{sec:method}

\begin{figure*}[!t]
    \centering
    \includegraphics[width=0.98\textwidth]{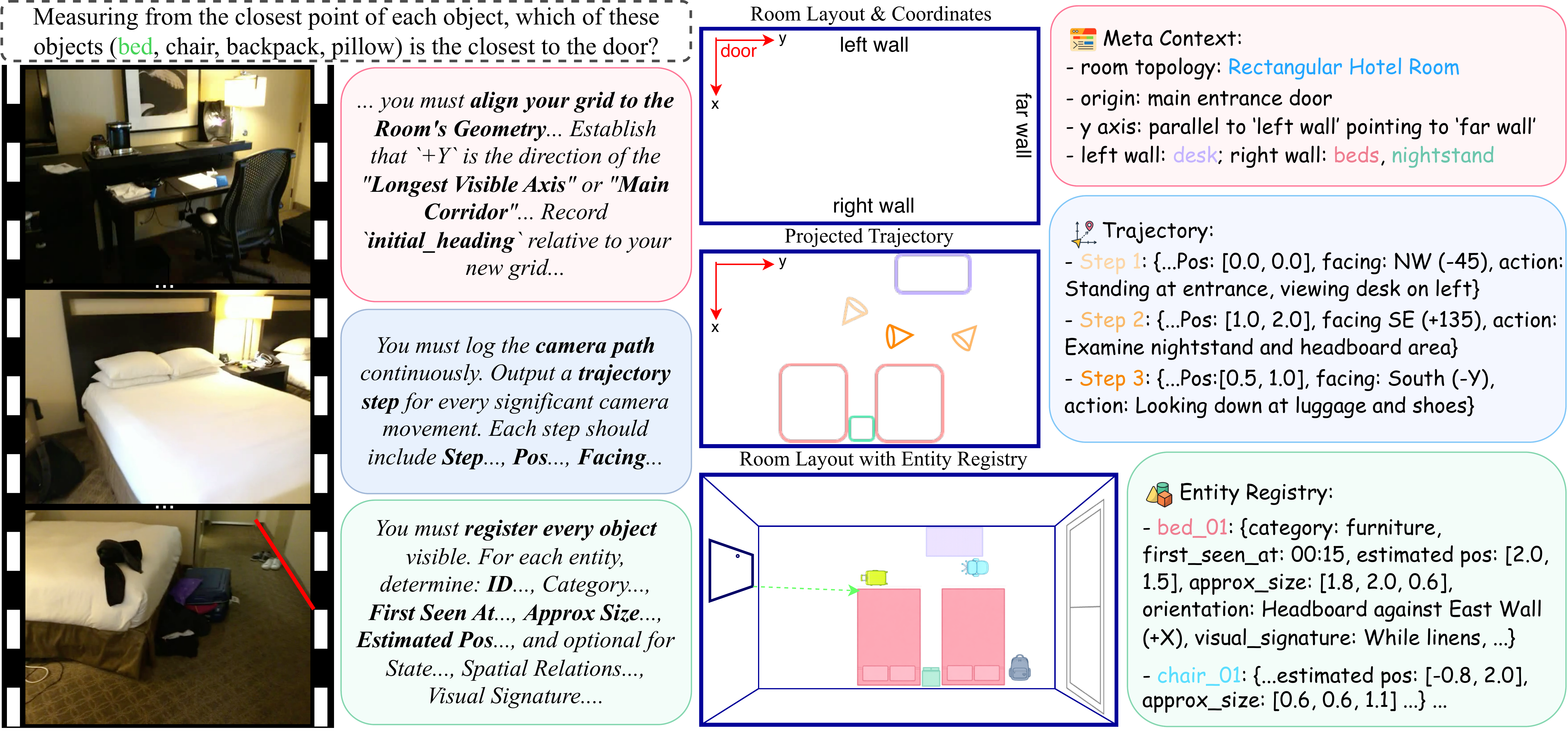}
    \caption{\textit{Illustration of our Textual Representation of Allocentric Context from Egocentric Video (TRACE).} We construct TRACE by aligning a global coordinate system with the room layout and geometry, logging the camera trajectory across temporal steps, and registering visible objects with key attributes, estimated positions, and spatial relations. Here, we also show the key prompts used to guide MLLMs to generate this intermediate reasoning trace.}
    \vspace{-6pt}
    \label{fig:method}
\end{figure*}

Standard prompting methods, such as Chain-of-Thought (CoT)~\cite{wei2022chainofthought}, encourage Multimodal Large Language Models (MLLMs) to generate intermediate reasoning steps to bridge the gap between input and output. While effective for arithmetic and symbolic tasks~\cite{cobbe2021training,hudson2019gqa}, standard chain of thought and other linguistic prompting strategies often fall short or even hurt performance on complex spatial reasoning tasks~\cite{yang2025thinking}. Our key intuition is that MLLMs may need to explicitly reason over an intermediate global representation of the 3D scene to complement the egocentric video inputs used in most spatial intelligence benchmarks.

To this end, drawing inspiration from human cognitive processes \cite{marr1978representation}, we introduce \textbf{Textual Representation of Allocentric Context from Egocentric Video} (\textbf{TRACE}), a method that encourages MLLMs to generate a text-based allocentric representation of the 3D environment that facilitates spatial question answering. In the following sections, we first introduce the problem setting of spatial question answering with prompting, then describe the key components of our TRACE design, and finally elaborate on the inference schema.

\subsection{Problem Formulation}

We formulate spatial reasoning as a generation task conditioned on a given egocentric video $V = \{v_1, ..., v_T\}$ and a natural language query $Q$, with the objective of generating the answer $A$.

Standard CoT approaches model the probability $\mathrm{P}(A, R|V, Q)$, where $R$ is a specified reasoning trace. However, previous reasoning traces often fail to capture the geometric structure required for spatial tasks~\cite{yang2025mmsi}. We instead enforce a protocol where the reasoning trace takes the form of a Textual Representation of Allocentric Context from Egocentric Video, denoted as $\mathcal{G}$. The inference process is formalized as a single-turn generation maximizing:
$$\hat{A}, \hat{\mathcal{G}} = \operatorname*{argmax}_{A, \mathcal{G}} \underbrace{P(A | \mathcal{G}, V, Q)}_{\text{Reasoning Parser}} \cdot \underbrace{P(\mathcal{G} | V, Q)}_{\text{Spatial Descriptor}}$$
Here, the Spatial Descriptor produces intermediate reasoning steps as TRACE, which the Reasoning Parser then uses to generate the final answer.

\subsection{Key Components of TRACE}

We formally define TRACE as a tuple $\mathcal{G} = \langle \mathcal{M, \mathcal{T}, \mathcal{E}}\rangle$. Here, $\mathcal{M}$ denotes the meta context, including the room topology, grid alignment, and the observer's initial heading. The trajectory $\mathcal{T} = \{(t_k, p_k, \phi_k)\}^{K}_{k=0}$ records the observer's position $p_k \in \mathbb{R}^d$ and heading $\phi_k$ at discrete time steps $t_k$. Finally, $\mathcal{E} = \{e_j\}^{N}_{j=1}$ denotes the set of $N$ entities.

\paragraph{Meta Context}
A common failure mode in spatial reasoning arises from losing track of camera initialization and the corresponding coordinate system. We propose a Room Aligned Coordinate System that is initialized from a coarse room layout sketch, for example a rectangular bedroom. We fix the origin $[0, 0]$ at starting position of the observer, and then establishes the y axis by detecting the most salient straight line characterized by static large objects rather than the camera's initial heading.

\paragraph{Camera Trajectory}
Static maps fail to capture the dynamic nature of video. To address this limitation, we require the model to reconstruct the observer path as a discrete sequence of steps using the established coordinate system and large static objects from the Meta Context $\mathcal{M}$ as reference points. For each step, TRACE records the timestamp, estimated position $[x, y]$, and the camera's facing direction. We approximate camera direction using eight discrete orientations defined by the cardinal directions, with the y axis aligned with north, as accurate numerical angle estimation is difficult for the Scene Descriptor and continuous pose representations pose challenges for the Reasoning Parser. In addition, we include an action property that encodes the camera centric motion context. Our formulation thus effectively reconstructs the surveyor's path, allowing the model to answer navigation and route-planning questions by traversing the generated static map rather than relying on transient visual memory.

\paragraph{Entity Registry}
Instead of predicting loose grid cells as in the Cognitive Map~\cite{yang2025thinking}, our model maintains a registry of observed entities with detailed attributes throughout the temporal sequence. To prevent object duplication and ensure precise localization, we enforce a structured schema for each object entity:
\begin{itemize}
    \item \textit{Temporal Stamping:} Each entity $e_i$ must include a timestamp recording its first seen time, aiding in object tracking.
    \item \textit{Visual Signature:} Each entity includes a brief appearance based description that captures its salient visual attributes, which helps disambiguate visually similar instances across time.
    \item \textit{Metric Estimation:} TRACE records plausible 2D coordinates $[x, y]$ in meters for every entity relative to the grid origin. While these coordinates are estimates, the act of estimation forces the model to resolve spatial relations (e.g., \textit{near}, \textit{between}) into geometric constraints.
    \item \textit{Spatial Relations:} Each entity records its relative spatial relations to nearby entities using natural language, providing complementary relational cues beyond absolute coordinates.
    \item \textit{Strict Serialization:} Entities should be listed individually (e.g., \texttt{chair\_01}, \texttt{chair\_02}) rather than grouped, ensuring granular counting and positional accuracy.
\end{itemize}

\subsection{Inference Mechanism}

The inference of our standard implementation is performed in a single pass. We condition the generation process to explicitly yield the schema-compliant representation~$\mathcal{G}$ prior to the final response. This acts as a structured Chain-of-Thought, where the generation~$\mathcal{G}$ effectively loads the context window with a ``spatial cache'' of the environment. The final answer is then derived via TRACE-conditioned inference, which jointly accounts for the egocentric video input and queries the cached TRACE to compute Euclidean distances between objects coordinates~$\mathcal{E}$ or traverse nodes in~$\mathcal{T}$. This mechanism improves final answer accuracy by grounding answer generation in previously generated and verifiable geometric constraints.

\section{Experiments}
\setlength{\textfloatsep}{10pt plus 1pt minus 2pt}
\setlength{\dbltextfloatsep}{10pt plus 1pt minus 2pt}
\setlength{\floatsep}{8pt plus 1pt minus 2pt}
\setlength{\dblfloatsep}{8pt plus 1pt minus 2pt}
\setlength{\intextsep}{8pt plus 1pt minus 2pt}

\definecolor{oursgreen}{RGB}{220,245,220}
\newcolumntype{L}[1]{>{\hsize=#1\hsize\raggedright\arraybackslash}X}
\newcolumntype{C}[1]{>{\hsize=#1\hsize\centering\arraybackslash}X}

\begin{table*}[t]
\captionsetup{skip=6pt}
\caption{\textit{Evaluation results on the VSI benchmark.} We report average performance and detailed breakdowns across numerical-answer and multiple-choice tasks, under proprietary and open-sourced base models. Best results are in \textbf{bold}, and second-best are \underline{underlined}.}
\label{tab:vsi}
\centering
\setlength{\tabcolsep}{3.5pt}
\scriptsize
\renewcommand{\arraystretch}{1.0}
\begin{tabularx}{0.98\linewidth}{@{} L{1.2} C{1.1} *{8}{C{0.96}} @{}}
\toprule
\multirow{2}{*}{Methods} &
\multirow{2}{*}{Avg.} &
\multicolumn{4}{c}{\cellcolor{yellow!35}Numerical Answer} &
\multicolumn{4}{c@{}}{\cellcolor{cyan!25}Multiple-Choice Answer} \\
\cmidrule(r){3-6} \cmidrule(l){7-10}
& &
\cellcolor{yellow!12}\makebox[0pt][c]{Obj.\ Cnt.} &
\cellcolor{yellow!12}\makebox[0pt][c]{Abs.\ Dist.} &
\cellcolor{yellow!12}\makebox[0pt][c]{Obj.\ Size} &
\cellcolor{yellow!12}\makebox[0pt][c]{Room Size} &
\cellcolor{cyan!10}\makebox[0pt][c]{Rel.\ Dist.} &
\cellcolor{cyan!10}\makebox[0pt][c]{Rel.\ Dir.} &
\cellcolor{cyan!10}\makebox[0pt][c]{Route} &
\cellcolor{cyan!10}\makebox[0pt][c]{Order} \\
\midrule
\multicolumn{10}{@{}l@{}}{\textit{Gemini 3 Pro as base model}} \\
Direct & 52.61 & 33.77 & 32.57 & 67.09 & 42.99 & 62.54 & 50.52 & 51.03 & 70.71 \\
CoT    & 53.65 & 30.35 & 34.54 & 64.05 & 40.76 & 61.78 & 58.09 & \textbf{61.34} & 71.96 \\
ToT    & 58.88 & 44.55 & \textbf{42.12} & 72.20 & 45.55 & \underline{65.35} & 57.83 & 55.62 & \textbf{73.73} \\
LtM    & 59.52 & 45.19 & 40.72 & \underline{73.36} & 44.15 & \textbf{65.82} & \underline{60.40} & 53.59 & \underline{73.64} \\
CM     & \underline{59.72} & \underline{46.70} & \underline{41.43} & 72.49 & \textbf{50.14} & 63.69 & 58.62 & 55.50 & 72.61 \\
\rowcolor{oursgreen}
Ours   & \textbf{60.15} & \textbf{47.55} & 38.82 & \textbf{73.90} & \underline{45.62} & 63.85 & \textbf{61.70} & \underline{58.01} & 72.97 \\
\midrule
\multicolumn{10}{@{}l@{}}{\textit{Qwen2.5-VL-72B-Instruct as base model}} \\
Direct & 36.28 & \textbf{33.36} & 20.53 & 49.31 & \underline{41.49} & \textbf{43.38} & 27.79 & 32.47 & \textbf{44.01} \\
CoT    & 29.78 & 21.27 & 24.95 & 16.31 & 40.94 & 39.44 & 33.16 & 28.87 & \underline{43.53} \\
ToT    & \underline{38.06} & 17.89 & 26.20 & 53.15 & \textbf{47.01} & 41.55 & \underline{36.78} & \textbf{35.05} & \textbf{44.01} \\
LtM    & 38.01 & \underline{23.27} & \textbf{31.39} & \underline{54.49} & 38.68 & \underline{42.96} & 34.71 & 29.90 & 36.73 \\
CM     & 35.47 & 21.58 & 15.67 & 52.65 & 37.26 & 39.44 & 36.05 & \underline{34.54} & 42.39 \\
\rowcolor{oursgreen}
Ours   & \textbf{39.38} & 22.05 & \underline{28.03} & \textbf{59.98} & 38.99 & 40.85 & \textbf{37.40} & 31.96 & 42.56 \\
\midrule
\multicolumn{10}{@{}l@{}}{\textit{MiMo-VL-7B as base model}} \\
Direct & \underline{39.79} & \textbf{36.02} & 29.84 & 52.38 & 42.95 & 40.14 & \underline{33.78} & 31.44 & 47.41 \\
CoT    & 37.49 & 34.27 & 23.50 & 48.52 & \underline{43.23} & 38.73 & 32.75 & 27.84 & 49.23 \\
ToT    & 39.14 & 29.45 & \underline{30.44} & \underline{54.26} & 40.14 & \underline{41.41} & 32.02 & \underline{32.47} & 46.60 \\
LtM    & 38.34 & \underline{35.09} & 24.47 & 48.22 & \textbf{44.48} & \textbf{43.10} & 30.79 & \textbf{35.05} & \underline{49.50} \\
CM     & 36.85 & 27.43 & 23.14 & 50.14 & 39.06 & \underline{41.41} & 32.54 & 27.84 & 46.76 \\
\rowcolor{oursgreen}
Ours   & \textbf{41.42} & 33.27 & \textbf{31.51} & \textbf{58.67} & 41.56 & 39.44 & \textbf{35.33} & 28.87 & \textbf{51.29} \\
\bottomrule
\end{tabularx}
\end{table*}

\begin{table*}[t]
  \centering
  \captionsetup{skip=6pt}
  \scriptsize
  \caption{\textit{Evaluation results on the OST benchmark.} Results are reported across agent state understanding, visible information reasoning, and agent–object spatial relationship tasks, under proprietary and open-sourced base models. Best results are in \textbf{bold}, and second-best are \underline{underlined}.}
  \label{tab:ost}
  \setlength{\tabcolsep}{3 pt}
  \renewcommand{\arraystretch}{1.0}
  \begin{tabularx}{0.98\linewidth}{@{} L{1.2} C{1.1} *{15}{C{0.98}} @{}}
    \toprule
    \multirow{3}{*}{Methods} &
    \multirow{3}{*}{Avg.} &
    \multicolumn{5}{c}{\cellcolor{yellow!35}Agent Visible Info} &
    \multicolumn{6}{c}{\cellcolor{cyan!25}Agent-object Spatial Relationship} &
    \multicolumn{4}{c@{}}{\cellcolor{orange!35}Agent State} \\
    \cmidrule(lr){3-7} \cmidrule(lr){8-13} \cmidrule(l){14-17}
    & 
    & 
    \multicolumn{2}{c}{\cellcolor{yellow!20}Exist.} &
    \cellcolor{yellow!20}Quant. &
    \cellcolor{yellow!20}Divers. &
    \cellcolor{yellow!20}Order &
    \multicolumn{3}{c}{\cellcolor{cyan!15}Direct.} &
    \multicolumn{3}{c}{\cellcolor{cyan!15}Dist.} &
    \multicolumn{2}{c}{\cellcolor{orange!20}Pos.} &
    \multicolumn{2}{c@{}}{\cellcolor{orange!20}Orient.}\\
    & 
    & 
    \cellcolor{yellow!12}JUD. & 
    \cellcolor{yellow!12}TEMP. & 
    \cellcolor{yellow!12}CNT. & 
    \cellcolor{yellow!12}JUD. & 
    \cellcolor{yellow!12}JUD. & 
    \cellcolor{cyan!10}JUD. & 
    \cellcolor{cyan!10}TEMP. & 
    \cellcolor{cyan!10}EST. & 
    \cellcolor{cyan!10}JUD. & 
    \cellcolor{cyan!10}TEMP. & 
    \cellcolor{cyan!10}EST. &
    \cellcolor{orange!12}JUD. &
    \cellcolor{orange!12}EST. & 
    \cellcolor{orange!12}JUD. & 
    \cellcolor{orange!12}EST. \\
    \midrule
    \multicolumn{17}{@{}l@{}}{\textit{Gemini 3 Pro as base model}} \\
    Direct & 59.22 & \textbf{96.72} & 84.87 & \textbf{68.75} & 89.66 & \textbf{82.54} & 54.27 & 48.15 & 28.63 & \textbf{60.61} & \underline{54.55} & 30.00 & \underline{71.43} & 22.78 & \underline{72.60} & 22.68 \\
    CoT    & 59.26 & 82.24 & \textbf{93.99} & 65.96 & \underline{96.55} & 77.78 & 52.24 & \textbf{62.96} & 27.84 & 52.76 & 50.00 & \underline{31.64} & \underline{71.43} & 24.26 & \underline{72.60} & 26.67 \\
    ToT    & 59.20 & 94.54 & 83.55 & \underline{66.67} & 93.10 & \underline{80.65} & 54.27 & \underline{54.39} & \underline{31.00} & 55.61 & 53.85 & \textbf{32.36} & 61.76 & 29.07 & \underline{72.60} & 24.58 \\
    LtM    & \underline{59.27} & \underline{95.65} & 85.52 & \underline{66.67} & \textbf{100.0} & 76.19 & 53.00 & 50.91 & 25.88 & 55.78 & 53.03 & 31.23 & \underline{71.43} & \textbf{35.85} & 69.86 & 18.06 \\
    CM     & 59.04 & 95.05 & 86.18 & \textbf{68.75} & 89.66 & 77.78 & \underline{55.72} & 48.15 & 25.40 & \underline{60.30} & 49.23 & 26.30 & \textbf{80.00} & 24.81 & 69.86 & \underline{28.47} \\
    \rowcolor{oursgreen}
    Ours   & \textbf{60.42} & 95.05 & \underline{86.58} & \underline{66.67} & 96.43 & 77.78 & \textbf{57.00} & 52.73 & \textbf{34.12} & \underline{60.30} & \textbf{56.45} & 31.25 & 54.29 & \underline{31.92} & \textbf{76.71} & \textbf{29.03} \\
    \midrule
    \multicolumn{17}{@{}l@{}}{\textit{MiMo-VL-7B as base model}} \\
    Direct     & 62.65 & \textbf{92.39} & 51.63 & 53.06 & \textbf{100.0} & 85.71 & 63.68 & \underline{21.05} & \underline{34.12} & 76.38 & \underline{40.91} & \textbf{28.77} & \textbf{100.0} & 9.26 & 91.78 & \underline{40.28} \\
    CoT        & 61.69 & 89.67 & 48.37 & 48.98 & \textbf{100.0} & 82.54 & 68.66 & 15.79 & 33.33 & 75.38 & 39.39 & \textbf{28.77} & \underline{97.14} & 11.30 & 89.04 & 39.31 \\
    ToT        & 62.20 & 91.30 & 52.29 & 51.02 & \textbf{100.0} & \textbf{90.48} & 64.68 & 15.79 & 29.22 & 75.38 & \underline{40.91} & \underline{27.53} & \textbf{100.0} & \underline{15.74} & 84.93 & \textbf{41.39} \\
    LtM        & 63.75 & 88.04 & 44.44 & \textbf{63.27} & \textbf{100.0} & \underline{88.89} & \textbf{77.11} & \underline{21.05} & \textbf{35.88} & \underline{76.88} & \underline{40.91} & 22.33 & \textbf{100.0} & 7.78 & \textbf{100.0} & 36.94 \\
    CM         & \underline{64.00} & 88.59 & \underline{54.90} & 57.14 & \textbf{100.0} & \underline{88.89} & \underline{69.15} & \underline{21.05} & 33.92 & \textbf{78.89} & 36.36 & \underline{27.53} & \textbf{100.0} & 11.85 & \underline{97.26} & 38.89 \\
    \rowcolor{oursgreen}
    Ours       & \textbf{65.04} & \underline{91.85} & \textbf{57.52} & \underline{61.22} & \textbf{100.0} & 87.30 & \underline{69.15} & \textbf{24.56} & 32.35 & 74.87 & \textbf{42.42} & 26.44 & \textbf{100.0} & \textbf{29.07} & 94.52 & 38.06 \\
    \bottomrule
  \end{tabularx}
\end{table*}

\subsection{Experimental Setup}

\paragraph{Benchmarks} We consider two spatial intelligence related benchmarks: VSI-Bench~\cite{yang2025thinking} and OST-Bench~\cite{lin2025ostbench}.

VSI-Bench is a video-based benchmark built from egocentric indoor scene scans, containing 5{,}130 question-answer (QA) pairs across 288 real-world videos. It covers eight tasks spanning configurational, measurement-estimation, and spatiotemporal reasoning. In contrast, OST-Bench assesses online spatio-temporal understanding from the perspective of an embodied agent actively exploring a scene. Comprising 1{,}386 scenes and 10{,}165 QA pairs, it employs a multi-round dialogue format that requires models to process incrementally acquired observations and integrate historical memory to answer questions regarding the agent's state, visible information, and spatial relationships. In this work, we evaluate on the full set of VSI-Bench, while for OST-Bench, we use a reproducible random subset consisting of 200 scenes and 1{,}396 QA pairs.

\paragraph{Metrics} Current Spatial AI benchmarks mainly follow two formats: multiple-choice questions (MCQ) and numerical questions. For MCQ, we report Accuracy (Acc). To evaluate model predictions, we extract the answer option using exact matching, supplemented by fuzzy matching to robustly handle variations in model output formats (e.g., capturing the option letter or full text).

For numerical questions, we adopt the Mean Relative Accuracy (MRA) introduced by~\citet{yang2025thinking}. MRA quantifies the proximity of a predicted value $\hat{y}$ to the ground truth $y$ by averaging performance across a range of strictness thresholds $\mathcal{C} = \{0.5, 0.55, \dots, 0.95\}$. MRA is formally defined as:
$$
    \text{MRA} = \frac{1}{|\mathcal{C}|}\sum_{\theta \in \mathcal{C}}\mathbb{I}\left(\frac{|\hat{y} - y|}{y} < 1 - \theta\right)
$$
where $\mathbb{I}(\cdot)$ denotes the indicator function. A prediction is considered correct at threshold $\theta$ only if its relative error is less than $1 - \theta$.

\paragraph{Model Selection} We validate the effectiveness of our approach using Gemini 3 Pro~\cite{geminiteam2025geminifamilyhighlycapable} as our primary proprietary model. All open-source baselines are evaluated using their default configurations and parameters. For VSI-Bench, we report main results using both Qwen2.5-VL-72B~\cite{Qwen2.5-VL} and MiMo-VL-7B~\cite{coreteam2025mimovltechnicalreport}. Additional experiments on VSI-Bench with other state-of-the-art models, including o3~\cite{openai2025o3} and GLM-4.5V~\cite{vteam2025glm45vglm41vthinkingversatilemultimodal}, are detailed in Appendix~\ref{app:evaluation}. For OST-Bench, we adopt MiMo-VL-7B as our open-source backbone, omitting the Qwen series due to its documented limitations in multi-turn instruction-following settings~\cite{lee2025multiverse}.

\subsection{Experimental Results}
\label{sec:exp}

\paragraph{Comparison of Different Prompting Methods}
We first contrast our method with previously proposed prompting methods that have demonstrated effectiveness on general VQA tasks. Specifically, we consider the following prompting strategies:

\begin{itemize} 
\item \textit{Chain-of-Thought (CoT)}~\cite{wei2022chainofthought}: Elicits a step-by-step reasoning trace to bridge the gap between the input and the final answer.
\item \textit{Tree-of-Thought (ToT)}~\cite{yao2023treeofthought}: Explores a tree of potential reasoning paths, evaluating and selecting the most promising intermediate thoughts to derive the answer.
\item \textit{Least-to-Most (LtM)}~\cite{zhou2023leasttomost}: Decomposes complex spatial queries into manageable sub-problems, solving them sequentially to guide the final inference.
\item \textit{Cognitive Map (CM)}~\cite{yang2025thinking}: Instructs the model to construct a 10×10 semantic grid capturing the coarse layout of relevant objects before answering. \end{itemize}

To ensure fair comparison and seamless integration of different prompting techniques, we keep the prompting scaffold the same (e.g., identical input formatting, answer constraints, and post-processing), and vary only the method-specific instructions required by each prompting technique. We provide all prompts in Appendix~\ref{app:prompt}.

We evaluate our method, TRACE, alongside the Direct baseline and prior prompting strategies. Results are summarized in Tab.~\ref{tab:vsi} and Tab.~\ref{tab:ost}.

\begin{figure}
    \centering
    \includegraphics[width=0.84\columnwidth]{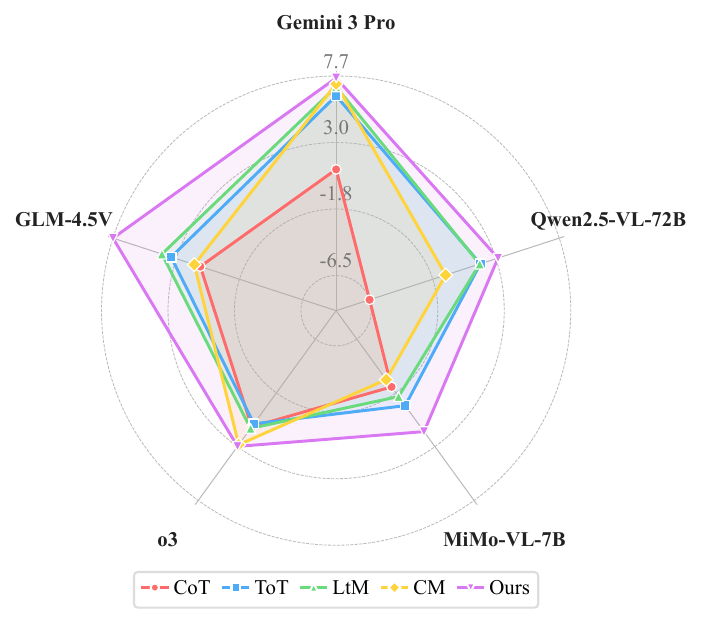}
    \caption{\textit{Performance gains across models on VSI-Bench.} TRACE yields consistent, state-of-the-art performance gains compared to Direct prompting baselines, across various model architectures and parameter scales.}
    \label{fig:perf_gain}
\end{figure}

On VSI-Bench, advanced prompting methods consistently improve performance for Gemini, but yield only marginal gains or even compromise performance for Qwen. This discrepancy is likely due to the weaker instruction-following capability of the Qwen series, which limits its ability to effectively leverage prompting strategies for in-depth reasoning. Notably, our proposed TRACE yields substantial performance improvements of +7.54\%, +3.10\% and +1.63\% for Gemini, Qwen and MiMo, respectively. These results demonstrate the robustness of our approach across different base models. In addition, we note that the latest Gemini 3 series incorporate step-by-step thinking instruction during training data construction, which likely leads to stronger alignment with existing prompting strategies and thus an inherent advantage. Even so, TRACE consistently outperforms these approaches on Gemini. Furthermore, additional experiments with other state-of-the-art models also demonstrate consistent performance gains with TRACE, as visualized in Fig~\ref{fig:perf_gain}.

On the OST benchmark, existing prompting strategies yield only marginal performance gains for both Gemini and MiMo models. This is because OST primarily evaluates multi-turn spatial reasoning, where step-by-step thinking prompts may hinder the model’s ability to accurately ground and update spatial context across turns. In contrast, TRACE yields a +1.2\% absolute performance gain on Gemini, and a +2.4\% gain on the open-source MiMo. Notably, for the compact MiMo backbone, spatial specific prompting (CM and TRACE) prove superior to general linguistic reasoning (CoT, LtM and ToT), underscoring the effectiveness of explicit geometric grounding for smaller models. We do acknowledge, however, that TRACE can lead to a performance drop in certain agent state predictions. This limitation arises because TRACE is currently formulated as a static global allocentric representation. While this global perspective provides a highly stable environment model for relational reasoning, it creates a decoupling from the rapid, dynamic egocentric updates required for precise real-time agent state tracking.

\begin{table*}[htbp]
\centering
\captionsetup{skip=6pt}
\caption{\textit{Systematic studies of different prediction settings for utilizing our proposed text-based spatial representation.} Among all settings, one-stage prompting yields the best performance on both Gemini-3 Pro and Qwen2.5-VL-72B.}
\label{tab:stage}
\scriptsize
\setlength{\tabcolsep}{2.3pt}
\renewcommand{\arraystretch}{0.95}
\begin{tabularx}{0.98\linewidth}{@{} L{1.55} C{0.75} *{8}{C{0.96}} @{}}
\toprule
{Input Setting} & {Avg.}
& \multicolumn{4}{c}{Numerical Answer}
& \multicolumn{4}{c}{Multiple-Choice Answer} \\
\cmidrule(r){3-6} \cmidrule(l){7-10}

& & 
Obj. Cnt. & 
Abs. Dist. & 
Obj. Size & 
Room Size & 
Rel. Dist. & 
Rel. Dir. & 
Route & 
Order \\
\midrule

\multicolumn{10}{@{}l@{}}{\textit{Proprietary model as base}} \\
Video Direct & 52.61 & 33.77 & 32.57 & 67.09 & 42.99 & 62.54 & 50.52 & 51.03 & 70.71 \\
One-Stage    & 60.15 & 47.55 & 38.82 & 73.90 & 45.62 & 63.85 & 61.70 & 58.01 & 72.97 \\
Two-Stage    & 58.52 & 42.25 & 36.73 & 72.10 & 52.17 & 58.75 & 63.50 & 51.35 & 74.01 \\
Text-Only    & 52.27 & 28.52 & 32.66 & 67.28 & 48.02 & 49.58 & 62.66 & 52.43 & 64.93 \\
\midrule

\multicolumn{10}{@{}l@{}}{\textit{Open-sourced model as base}} \\
Video Direct & 37.58 & 32.58 & 24.51 & 55.26 & 39.13 & 41.13 & 28.93 & 31.44 & 43.20 \\
One-Stage    & 38.92 & 25.47 & 26.93 & 58.18 & 40.42 & 37.46 & 36.15 & 29.38 & 45.79 \\
Two-Stage    & 32.85 & 16.80 & 19.75 & 42.33 & 26.46 & 37.32 & 34.19 & 34.02 & 45.95 \\
Text-Only    & 31.11 & 12.83 & 21.74 & 39.71 & 23.51 & 37.04 & 33.16 & 32.99 & 40.13 \\
\bottomrule
\end{tabularx}
\end{table*}

\paragraph{Comparison of Different Prediction Setting} We further examine how carefully designed text-based video representations can improve 3D spatial understanding. In particular, we explore the following prediction settings through which MLLMs can leverage such text-based representations:
\begin{itemize}
    \item \textit{One-Stage Inference} is the setup discussed in Sec.~\ref{sec:method}, where the model generates TRACE and answers the question using both the representation and the video input in a single pass.
    \item \textit{Two-Stage Inference} first generates TRACE, which is then treated as additional context and fed into the MLLM, together with the video input, for final question answering.
    \item \textit{Text-Only Inference} first generates our proposed TRACE and then uses an LLM to answer the question based solely on TRACE.
\end{itemize}

For fair comparison, we adopt the same MLLM and prompt components to construct TRACE representations across all settings. As shown in Tab.~\ref{tab:stage}, the text-only approach achieves on-par performance with the direct video-based method using Gemini, suggesting that TRACE provides an informative summary of the video sequence. Another important finding is that, for both Qwen and Gemini, the one-stage prompting setting outperforms the two-stage setting. This suggests that not only is the resulting text-based representation beneficial, but the reasoning process involved in generating it also plays a critical role in enabling MLLMs to make accurate predictions.

\begin{table*}[htbp]
\centering
\captionsetup{skip=6pt}
\caption{\textit{Comparison with existing text-based spatial representations and ablation studies of our method.} We use Qwen2.5-VL-72B as base and adopt text-only inference for more direct comparison.}
\label{tab:ablation}
\scriptsize

\setlength{\tabcolsep}{1.8pt}
\renewcommand{\arraystretch}{0.95}
\begin{tabularx}{0.98\linewidth}{@{} L{1.75} C{0.75} *{8}{C{0.94}} @{}}
\toprule
{Input Setting} & {Avg.}
& \multicolumn{4}{c}{Numerical Answer}
& \multicolumn{4}{c}{Multiple-Choice Answer} \\
\cmidrule(r){3-6} \cmidrule(l){7-10}

& & Obj. Cnt. & Abs. Dist. & Obj. Size & Room Size & Rel. Dist. & Rel. Dir. & Route & Order \\
\midrule
Cognitive Map    & 21.41 & 7.82  & 9.86  & 8.17  & 21.22 & 33.94 & 35.23 & 32.99 & 30.26 \\
Spatial Caption  & 27.58 & 14.90 & 14.20 & 24.86 & 30.59 & 36.06 & 34.40 & 36.60 & 36.73 \\
\midrule
Ours             & 31.11 & 12.83 & 21.74 & 39.71 & 23.51 & 37.04 & 33.16 & 32.99 & 40.13 \\
Ours w/o Trajectory & 29.19 & 10.16 & 18.19 & 33.93 & 29.41 & 35.92 & 37.19 & 31.96 & 32.85 \\
Ours w/o Entity Registry & 25.87 & 6.11 & 28.84 & 7.41 & 19.69 & 41.69 & 37.50 & 30.41 & 33.50 \\
\bottomrule
\end{tabularx}
\end{table*}

\paragraph{Comparison with other Text-based Spatial Representations}

\begin{figure}[t]
    \centering
    \includegraphics[width=0.98\columnwidth]{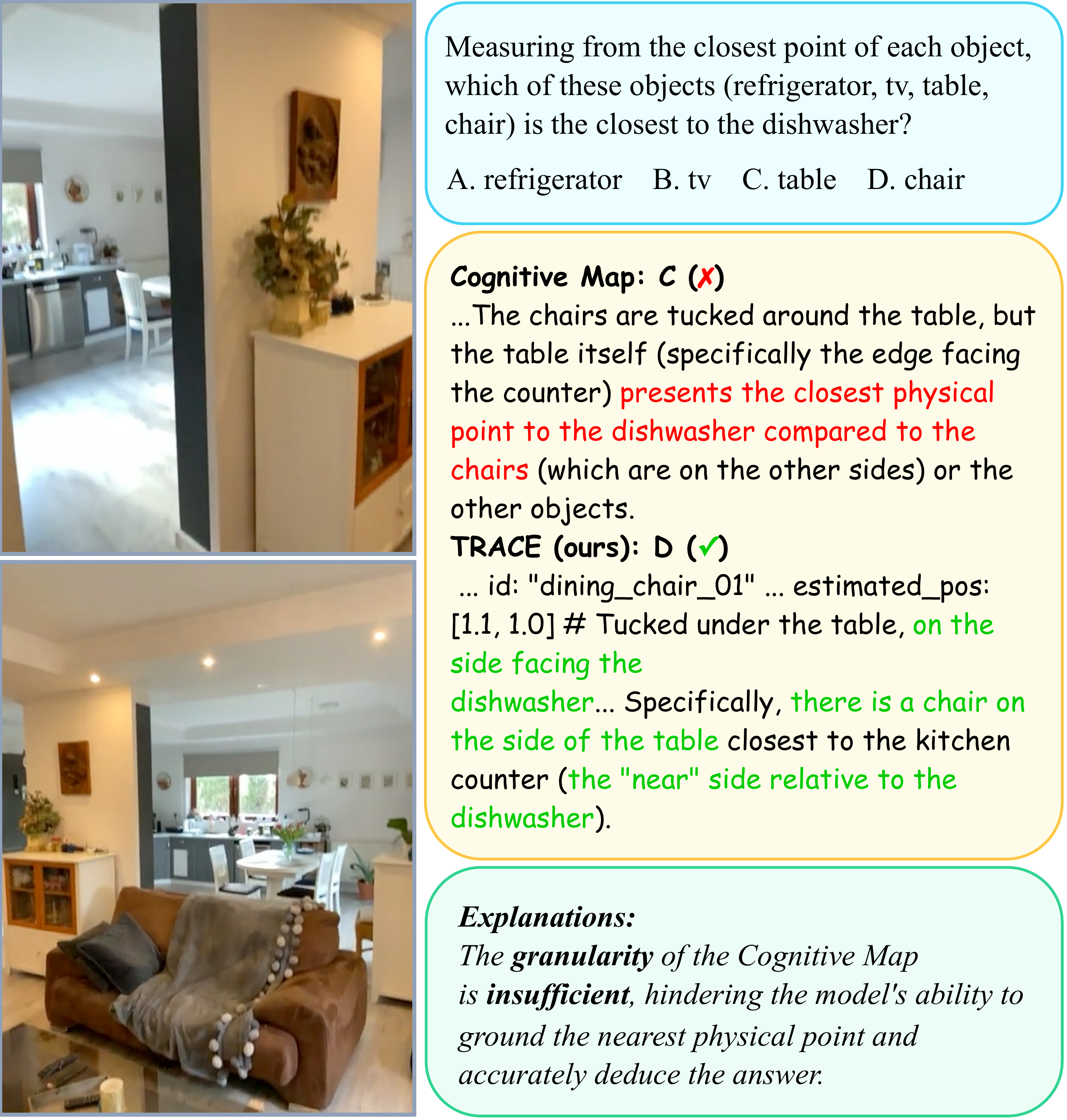}
    \captionsetup{skip=8pt}    
    \caption{\textit{A visual illustration demonstrates that TRACE is more effective than the cognitive map (CM) approach.} Notably, the CM lacks the 3D granularity required for many spatial reasoning tasks.}
    \label{fig:vis}
    \vspace{-3pt}
\end{figure}

We compare our method with the most relevant cognitive map representation proposed in~\cite{yang2025thinking} and a spatial captioning approach inspired by~\cite{zhang2024simple}, which sequentially describes the spatial components of the video sequence. To explicitly quantify the benefits of the text-based representation, we adopt the aforementioned text-only inference setting. As shown in Tab.~\ref{tab:ablation}, our method outperforms Cognitive Map by 9.7\% and Spatial Caption by 3.53\% on the VSI-Bench, highlighting the advantage of our proposed spatial representation. Furthermore, a visual comparison in Fig.~\ref{fig:vis} shows this advantage qualitatively, illustrating that TRACE captures the essential 3D granularity required for complex spatial reasoning that the cognitive map approach lacks.

\paragraph{Ablation Studies} We ablate the key components of our method in Tab.~\ref{tab:ablation}. Removing trajectory information results in a 1.92\% performance drop, while excluding entity registry leads to a larger drop of 5.24\%, suggesting camera trajectory and entity registry play important roles in spatial QA. As expected, removing the entity registry leads to a substantial performance drop on object related tasks, while removing camera trajectory mainly affects performance on distance and order related reasoning. In addition, we find that removing trajectory information improves performance on room size and relative direction tasks. This suggests that current MLLMs lack the ability to reliably estimate camera motion, which can confuse models on tasks that require alignment-based reasoning.

\subsection{Additional Analysis}

\begin{figure*}[t]
    \centering
    \begin{minipage}[t]{0.34\textwidth}
        \centering
        \includegraphics[width=\linewidth]{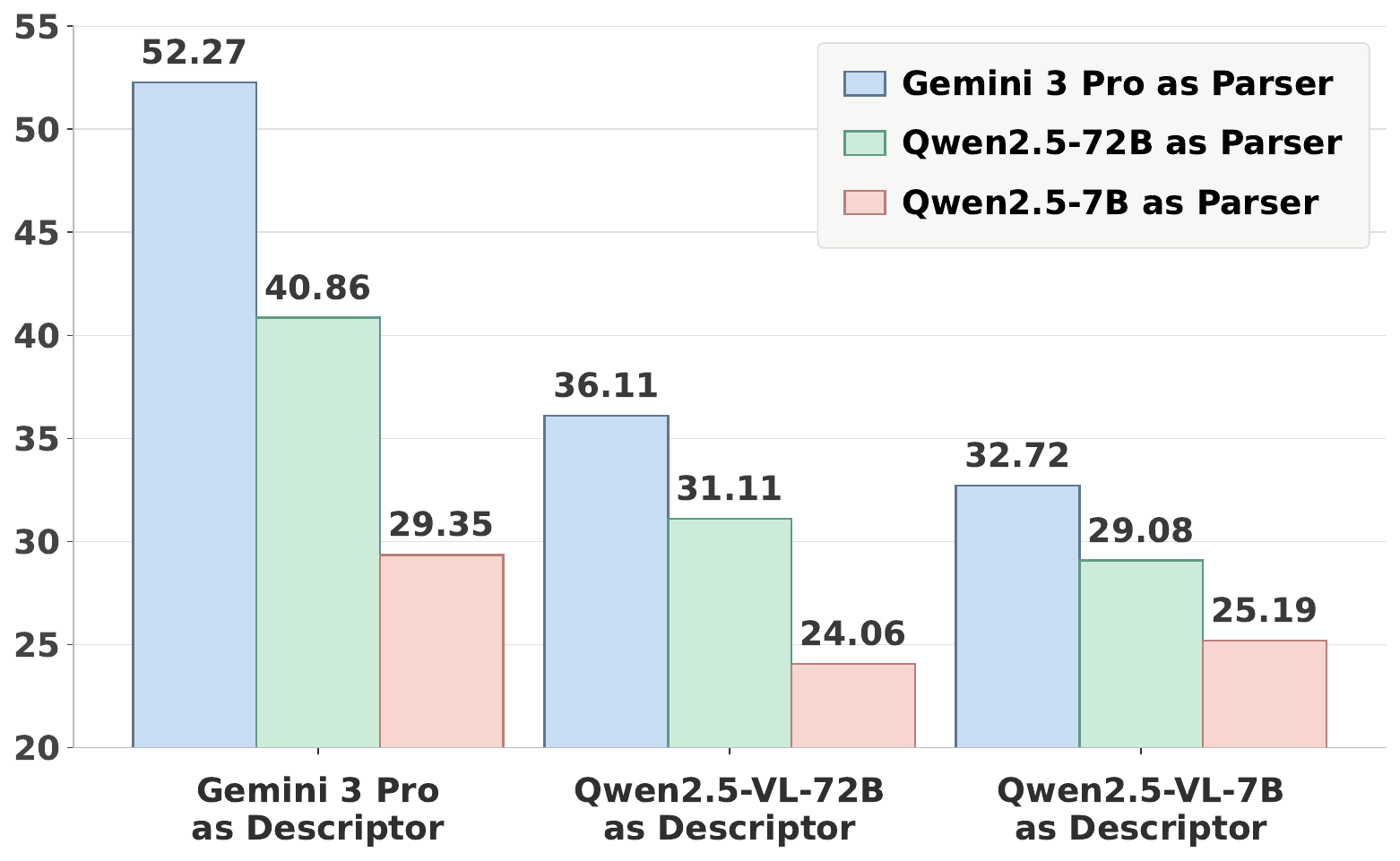}
        \caption{\textit{Decompositional analysis of the reasoning parser and spatial descriptor.} The Qwen series lags behind the state-of-the-art Gemini 3 on both spatial reasoning and visual perception.}
        \label{fig:decomposition}
    \end{minipage}
    \hfill
    \begin{minipage}[t]{0.64\textwidth}
        \centering
        \includegraphics[width=\linewidth]{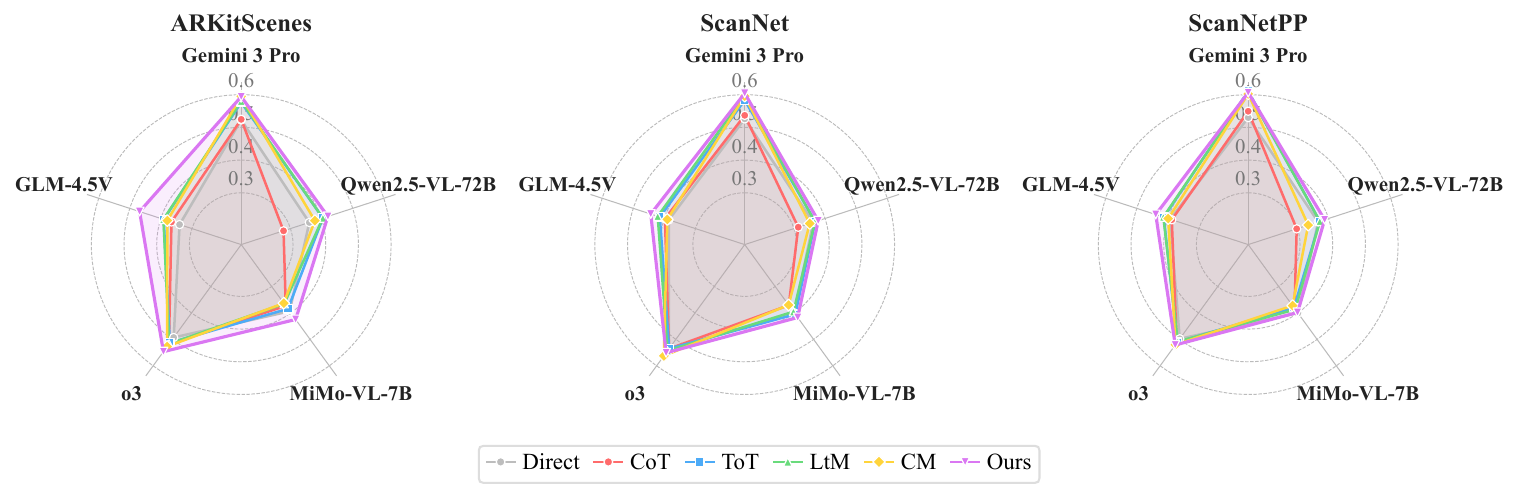}
        \caption{\textit{Stratified analysis on VSI-Bench.} Our method (TRACE) consistently achieves robust performance gains across all granular scene distributions, demonstrating reliable generalization across diverse spatial layouts and complexities without aliasing to specific environment types.}
        \label{fig:strat}
    \end{minipage}
    \vspace{-10pt}
\end{figure*}

\paragraph{Decomposing 3D Spatial Understanding}
Prior works~\cite{yang2025mmsi,yang2025thinking} have shown that existing MLLMs have limited 3D spatial understanding capabilities. We seek to provide an in-depth analysis of the underlying causes using a text only inference setting. Concretely, we decompose 3D reasoning into two stages: 3D visual perception and language-based spatial reasoning. Specifically, we use MLLMs as both a \emph{Spatial Descriptor} for 3D grounding and a \emph{Reasoning Parser} for spatial knowledge inference.

As shown in Fig.~\ref{fig:decomposition} using Gemini 3 Pro as a theoretical performance upper bound, we observe a significant performance drop when either the Descriptor or the Parser is replaced with Qwen2.5-VL-72B, especially when replacing the Descriptor. In addition, replacing the spatial descriptor from Qwen2.5-72B with Qwen2.5-7B results in only a marginal performance drop, whereas swapping the reasoning parser from Qwen2.5-72B to Qwen2.5-7B leads to a substantially larger degradation. This suggests that the two models exhibit comparable 3D visual perception capabilities, while the 72B variant has a markedly stronger reasoning capacity. Such decompositional analysis therefore helps identify the key bottlenecks in prevailing LLMs.
\vspace{-1pt}

\paragraph{Token Efficiency}
We observe that the token length induced by thinking-based methods is highly sensitive to the choice of model backbone. Notably, our method achieves greater token efficiency while delivering better performance than advanced baselines (e.g., ToT and LtM) on compact models like MiMo, underscoring its strong potential for lower-latency embodied AI deployments. The same trend holds for specific large models, such as GLM, although our method is slightly more token-intensive on some other large foundation models. We refer readers to Appendix~\ref{app:evaluation} for a more comprehensive breakdown and analysis. Importantly, optimizing token efficiency during the reasoning process constitutes a largely orthogonal research direction, which we leave for future work.
\vspace{-1pt}

\paragraph{Cross-Environment Generalization}
To investigate whether TRACE is biased toward specific environment types, we stratify the VSI-Bench results across its core underlying datasets: ARKitScenes~\cite{baruch2021arkitscenes}, ScanNet~\cite{dai2017scannet}, and ScanNetPP~\cite{yeshwanth2023scannet++}. These datasets represent a diverse range of indoor spatial layouts, scanning fidelities, and environmental complexities. As shown in Fig.~\ref{fig:strat}, our structured prompting approach consistently delivers robust performance gains across all three scene distributions and five different model architectures. This confirms that our method's effectiveness is not restricted to a specific environment type, but rather generalizes reliably across varied spatial features and complexities.
\setlength{\textfloatsep}{20pt plus 2pt minus 4pt}
\setlength{\dbltextfloatsep}{20pt plus 2pt minus 4pt}
\setlength{\floatsep}{12pt plus 2pt minus 2pt}
\setlength{\dblfloatsep}{12pt plus 2pt minus 2pt}
\setlength{\intextsep}{12pt plus 2pt minus 2pt}

\section{Conclusions}

We presented \textbf{TRACE}, a prompting approach that enables MLLMs to leverage the \textbf{T}extual \textbf{R}epresentation of \textbf{A}llocentric \textbf{C}ontext from \textbf{E}gocentric Video as an intermediate reasoning trace for spatial understanding. By explicitly modeling scene structure through meta-context, camera trajectory, and entity-level grounding, TRACE consistently improves performance on VSI-Bench and OST-Bench across diverse proprietary and open-source model backbones. Comparisons against prior linguistic prompting methods and other text-based spatial representations, together with detailed ablation studies, validate the effectiveness of our design choices. We further provide insights into how to effectively leverage text-based representations and present decompositional analyses that reveal common failure modes in MLLM spatial reasoning. More broadly, we hope TRACE can serve as a simple and widely applicable interface for studying structured spatial reasoning in off-the-shelf MLLMs. Overall, our results suggest a promising direction for advancing spatial reasoning in MLLMs and motivate further exploration of cognitively inspired representations. 

\section*{Limitations and Future Work}

Our work presents an initial attempt to design text-based representations that facilitate effective spatial reasoning in MLLMs. Nevertheless, our approach is still subject to several limitations. First, our current framework is formulated as a static allocentric representation. While this ensures global topological consistency, it somewhat creates a decoupling from the dynamic egocentric updates required for precise real-time agent state tracking in multi-turn settings. Furthermore, for the sake of fair comparison, our current implementation relies on the vision-language model itself to generate the representation. In practice, incorporating specialized visual expert models may further enhance the accuracy of the generated scene structures.

There are several promising directions for future work. A natural next step is to develop a dynamic streaming TRACE framework that incrementally updates the camera trajectory and entity registry as new observations arrive, allowing the model to maintain a persistent world model while recursively re-projecting the agent's pose within the map. Another important direction is to internalize TRACE-like reasoning into MLLMs through training, for example, by using TRACE to automatically construct high-quality spatial reasoning supervision for supervised finetuning, and potentially further improving the resulting policies with reinforcement learning. This could enable structured spatial representations to become part of the model’s native reasoning process rather than an external prompt artifact. It would also be valuable to integrate specialized 3D perception modules into TRACE to improve the fidelity of the intermediate representation while preserving the flexibility of language-based reasoning. Future work could additionally study representation compression and token efficiency, since compact yet faithful spatial traces would be especially important for low-latency embodied agents. Finally, extending TRACE beyond QA to tasks such as navigation, planning, and manipulation may help establish whether structured textual world models can serve as a more general interface between perception and reasoning in multimodal systems. 

\section*{Acknowledgments}

This work was supported in part by Shanghai Artificial Intelligence Laboratory, the Zhiyuan Scholar Program of the Beijing Municipal Science and Technology Commission (Z251100008125045), NSFC Grants, and a research grant from the ByteDance Seed Team.

\bibliography{ref}

\appendix

\section*{Appendix}



This appendix provides technical details of our prompting method (Appendix~\ref{app:prompt}) and additional experiments and results (Appendix~\ref{app:evaluation}).

\section{Prompting Details}
\label{app:prompt}

We evaluate multiple prompting methods on our benchmark. Most methods share the same \emph{base system prompt} and differ mainly in the \emph{user prompt}. See more details below.

\subsection{Overall Structure}
\label{app:prompt:common}

\paragraph{Base system prompt.}
All methods (except the cognitive map method in \S\ref{app:prompt:linguistic}) use the following base system prompt. This design allows us to control variables across prompting strategies: although the Cognitive Map and our TRACE method require specialized system instructions to define their intermediate representations, we keep these system prompts as close as possible to the base prompt (e.g., identical task framing and answer constraints) so that observed differences are primarily attributable to the prompting protocol rather than unrelated changes in instruction wording.
\begin{quote}
\small
\texttt{SYSTEM\_PROMPT = """You are a multimodal large language model being evaluated on visual-spatial reasoning tasks with egocentric indoor videos.\\
\\
You are given:\\
- an egocentric video of an indoor environment, and\\
- a question about that video.\\
\\
Your goal is to answer the question as accurately as possible.\\
\\
Answer format:\\
- \{POST\_PROMPT\}\\
- Do NOT add extra text on the final answer line (no units, no explanations).\\
"""}%
\end{quote}

\paragraph{Prompt assembly.}
Most benchmarks have two types of question: multiple-choice question and open-ended question with numerical answer needed. We instantiate post prompt according to the question type:
\begin{quote}
\small
POST\_PROMPT\_NA: Please answer the question using a single word or phrase enclosed in backticks.\\
POST\_PROMPT\_MCA: Answer with the option's letter from the given choices only, enclosed in backticks.
\end{quote}

Given a user-prompt template, we construct the final user message by concatenating the user prompt with the question block, separated by blank lines and explicit field headers. For open-ended questions, we use:
\begin{quote}
\small
prompt = USER\_PROMPT + ``\textbackslash n\textbackslash nQuestion:\textbackslash n'' + question + ``\textbackslash n''.
\end{quote}
For multiple-choice questions, we additionally append the options block:
\begin{quote}
\small
prompt = USER\_PROMPT + ``\textbackslash n\textbackslash nQuestion:\textbackslash n'' + question
+ ``\textbackslash n\textbackslash nOptions:\textbackslash n'' + options\_str + ``\textbackslash n''.
\end{quote}
In all methods, the final answer must appear on the last line in the format \mbox{Answer: `X'} to satisfy \textsc{POST\_PROMPT}. (See more implementations in below scripts.)

\subsection{User Prompts for Linguistic Reasoning Methods}
\label{app:prompt:linguistic}

We evaluate several advanced prompting strategies, including Chain-of-Thought (CoT)~\cite{wei2022chainofthought}, Tree-of-Thoughts (ToT)~\cite{yao2023treeofthought}, Least-to-Most~\cite{zhou2023leasttomost}, and Cognitive Map prompting~\cite{yang2025thinking}. Below we provide the user-prompt templates used for each method.

\paragraph{Direct prompting.}
Direct prompting instructs the model to solve the task internally while suppressing any explicit reasoning, and to output only the final answer in the required format.

\begin{promptbox}{Direct Prompt}
\textbf{[System Prompt]}\\
(Base system prompt; see \S\ref{app:prompt:common}.)\\
\smallskip

\textbf{[User Prompt]}\\
\textbf{Reasoning protocol:}\\
\hspace{0.5em}-- Read the question carefully.\\
\hspace{0.5em}-- You may think through the problem internally, but do \emph{NOT} show your reasoning.\\
\hspace{0.5em}-- Directly provide the final answer in the required format.\\
\smallskip

\textbf{Output format:}\\
\noindent Answer: \texttt{`X'}\\
\smallskip

Now follow this protocol to answer the question below.
\end{promptbox}

\paragraph{Chain-of-Thought (CoT) prompting.}
For CoT prompting~\cite{wei2022chainofthought}, we explicitly request a step-by-step natural-language explanation before emitting the final answer line. The user prompt enforce a two-part output: a \emph{Reasoning} block with the step-by-step explanation, followed by a \emph{Final answer} block containing exactly \mbox{Answer: `X`} on the last line.

\begin{promptbox}{Chain-of-Thought Prompt}
\textbf{[System Prompt]}\\
(Base system prompt; see \S\ref{app:prompt:common}.)\\
\smallskip

\textbf{[User Prompt]}\\
\textbf{Reasoning protocol:}\\
\noindent\hspace*{0.5em}-- First, think step by step about the visual scene and the spatial relationships involved.\\
\noindent\hspace*{0.5em}-- Explain your reasoning clearly in natural language.\\
\noindent\hspace*{0.5em}-- At the end, provide a single final answer line in the required format.\\
\smallskip

\textbf{Output format:}\\
\noindent Reasoning: \\
\noindent [\emph{step-by-step explanation}]\\
\noindent Final answer: \\
\noindent Answer: \texttt{`X'}\\
\smallskip

Now follow this protocol to answer the question below.
\end{promptbox}

\paragraph{Tree-of-Thoughts (ToT) prompting.}
For Tree-of-Thoughts~\cite{yao2023treeofthought}, we enforces a three-stage procedure: (1) generate three distinct reasoning branches (``Thought 1--3''), (2) compare and select the best branch under consistency and spatial-coherence checks, and (3) output the final answer based on the selected branch. The output format includes the three thoughts, an explicit evaluation section, the chosen best thought, and then the final answer line.

\begin{promptbox}{Tree-of-Thoughts Prompt}
\textbf{[System Prompt]}\\
(Base system prompt; see \S\ref{app:prompt:common}.)\\
\smallskip

\textbf{[User Prompt]}\\
\textbf{Reasoning protocol:}\\
\noindent\hspace*{0.5em}-- \textbf{Step 1:} Generate multiple reasoning branches (\emph{thoughts}).\\
\noindent\hspace*{1.5em}-- Propose 3 plausible reasoning paths about the question.\\
\noindent\hspace*{1.5em}-- Each path should be coherent and may use different assumptions about the spatial layout.\\
\noindent\hspace*{0.5em}-- \textbf{Step 2:} Evaluate and compare the thoughts.\\
\noindent\hspace*{1.5em}-- Check consistency with the video evidence, spatial coherence, and contradictions.\\
\noindent\hspace*{1.5em}-- Select the most reliable thought overall.\\
\noindent\hspace*{0.5em}-- \textbf{Step 3:} Produce the final answer using the best thought.\\
\noindent\hspace*{1.5em}-- Use the most reliable thought to derive a single final answer in the required format.\\
\smallskip

\textbf{Output format:}\\
\noindent Thought 1:\\
\noindent [\emph{reasoning path 1}]\\
\noindent Thought 2:\\
\noindent [\emph{reasoning path 2}]\\
\noindent Thought 3:\\
\noindent [\emph{reasoning path 3}]\\
\noindent Evaluation: \\
\noindent\hspace*{0.5em}-- Thought 1: \dots\\
\noindent\hspace*{0.5em}-- Thought 2: \dots\\
\noindent\hspace*{0.5em}-- Thought 3: \dots\\
\noindent Best thought: Thought \texttt{X} because \dots\\
\noindent Final answer: \\
\noindent Answer: \texttt{`X'}\\
\smallskip

Now follow this protocol to answer the question below.
\end{promptbox}

\paragraph{Least-to-Most prompting.}
For Least-to-Most prompting~\cite{zhou2023leasttomost}, we ask the model to decompose each question into ordered subproblems from easiest to hardest (e.g., object identification $\rightarrow$ local relations $\rightarrow$ global layout $\rightarrow$ final decision), solve them sequentially while reusing intermediate results, and finally provide a single answer line. The output format is constrained to three stages: decomposition, solving subproblems, and final answer.

\begin{promptbox}{Least-to-Most Prompt}
\textbf{[System Prompt]}\\
(Base system prompt; see \S\ref{app:prompt:common}.)\\
\smallskip

\textbf{[User Prompt]}\\
\textbf{Reasoning protocol:}\\
\noindent\hspace*{0.5em}-- \textbf{Step 1:} Decompose the problem from easiest to hardest.\\
\noindent\hspace*{1.5em}-- Identify what type of visual-spatial task this is (e.g., object count, absolute distance,relative distance, relative direction, object size, room size, route planning, appearance order).\\
\noindent\hspace*{1.5em}-- Break the question into several subproblems ordered from the easiest to the most difficult.\\
\noindent\hspace*{1.5em}-- Typical stages may include:\\
\noindent\hspace*{2.5em}-- identifying relevant objects and regions,\\
\noindent\hspace*{2.5em}-- understanding local spatial relations,\\
\noindent\hspace*{2.5em}-- integrating them into a global spatial layout,\\
\noindent\hspace*{2.5em}-- making the final decision.\\
\noindent\hspace*{0.5em}-- \textbf{Step 2:} Solve subproblems in order.\\
\noindent\hspace*{1.5em}-- For each subproblem:\\
\noindent\hspace*{2.5em}-- Briefly name the subproblem.\\
\noindent\hspace*{2.5em}-- Explain how you solve it using evidence from the video.\\
\noindent\hspace*{2.5em}-- Reuse and refine results from previous subproblems.\\
\noindent\hspace*{1.5em}-- Keep the spatial layout consistent across all steps.\\
\noindent\hspace*{0.5em}-- \textbf{Step 3:} Produce the final answer.\\
\noindent\hspace*{1.5em}-- Based on the solved subproblems, give a single final answer in the required format.\\
\smallskip

\textbf{Output format:}\\
\noindent Step 1: Problem decomposition\\
\noindent\hspace*{0.5em}-- Subproblem 1: \dots\\
\noindent\hspace*{0.5em}-- Subproblem 2: \dots\\
\noindent\hspace*{0.5em}-- Subproblem 3: \dots\\
\noindent Step 2: Solving subproblems\\
\noindent\hspace*{0.5em}-- Subproblem 1:\\
\noindent\hspace*{1.5em}-- Reasoning: \dots\\
\noindent\hspace*{0.5em}-- Subproblem 2:\\
\noindent\hspace*{1.5em}-- Reasoning: \dots\\
\noindent\hspace*{0.5em}-- Subproblem 3:\\
\noindent\hspace*{1.5em}-- Reasoning: \dots\\
\noindent Step 3: Final answer\\
\noindent\hspace{1.0em}Answer: \texttt{`X'}\\
\smallskip

Now follow this protocol to answer the question below.
\end{promptbox}

\paragraph{Cognitive Map prompting.}
For Cognitive Map prompting~\cite{yang2025thinking}, we use a specialized description prompt that asks the model to first produce a textual cognitive map: for a fixed set of indoor categories of interest (COI). For example, this is the categories for VSI-bench:
\begin{quote}
\small
ceiling light, trash can, bed, heater, closet, pillow, backpack, chair, refrigerator, tv, nightstand,
keyboard, computer tower, coat hanger, table, trash bin, whiteboard, monitor, sofa, clock,
computer mouse, radiator, telephone
\end{quote}
the model estimates the center location of each object instance on a $10\times 10$ grid and outputs a dictionary in strict JSON form:
\begin{quote}
\small
\{``category name'': [``(x\_1, y\_1)'', \ldots], \ldots\}.
\end{quote}
After generating this cognitive map, the model answers the question, still respecting the same final answer constraint \mbox{Answer: `X'}.

\begin{promptbox}{Cognitive Map Prompt}
\label{cogmap}
\textbf{[System Prompt]}\\
You are an expert specialized in solving spatial understanding questions using text descriptions of egocentric video sequences.

You are given:\\
\noindent\hspace*{0.5em}-- an egocentric video of an indoor environment,\\
\noindent\hspace*{0.5em}-- a question about that video.

You need to firstly generate a ``Cognitive Map'' derived from the egocentric video, and then answer the question as accurately as possible.\\
\smallskip

HOW TO GENERATE THE COGNITIVE MAP:

[Task]\\
This video captures an indoor scene. Your objective is to identify specific objects within the video, understand the spatial arrangement of the scene, and estimate the center point of each object, assuming the entire scene is represented by a 10x10 grid.

[Rule]\\
\noindent\hspace*{0.5em}1. We provide the categories to care about in this scene: \textsc{\{categories\_of\_interest\}}. Focus ONLY on these categories.\\
\noindent\hspace*{0.5em}2. Estimate the center location of each instance within the provided categories, assuming the entire scene is represented by a 10x10 grid.\\
\noindent\hspace*{0.5em}3. If a category contains multiple instances, include all of them.\\
\noindent\hspace*{0.5em}4. Each object’s estimated location should accurately reflect its real position in the scene, preserving the relative spatial relationships among all objects.\\

[Output]\\
Present the estimated center locations for each object as a list within a dictionary. \emph{STRICTLY} follow this JSON format:
\textsc{\{"category name": ["(x\_1, y\_1)", ...], ...\}}.\\
\smallskip

Answer format:\\
- \textsc{\{POST\_PROMPT\}}\\
- Do NOT add extra text on the final answer line (no units, no explanations).\\
\smallskip

\textbf{[User Prompt]}\\
\textbf{Reasoning protocol:}\\
\noindent\hspace*{0.5em}-- Read the question and analyze the visual content carefully.\\
\noindent\hspace*{0.5em}-- Generate cognitive map as required to help determine spatial relationships and approximate distances.\\
\noindent\hspace*{0.5em}-- You may think through the problem internally, but do \emph{not} show your reasoning.\\
\noindent\hspace*{0.5em}-- Directly provide the final answer in the required format.\\
\smallskip

\textbf{Output format:}\\
\noindent Final answer:\\
\noindent Answer: \texttt{`X'}\\
\smallskip

Now follow this protocol to answer the question below.
\end{promptbox}

\paragraph{Textual Representation of Allocentric Context from Egocentric Video prompting.}

We propose a new prompting method to elicit a structured intermediate representation before answering: the model first generates a structured Textual Representation of Allocentric Context from Egocentric Video that summarizes the room-aligned coordinate system, the camera trajectory, and an entity registry with timestamps and estimated positions. The textual representation must be a single YAML document with three top-level sections:
\texttt{Meta\_Context}, \texttt{Trajectory}, and \texttt{Entity\_Registry}. Below is a detailed description which is included in the system prompt, and a reference TRACE prompt.

\begin{quote}\small
\textbf{Meta\_Context (required keys).}\\
\texttt{Meta\_Context:}\\
\hspace*{1.0em}\texttt{room\_topology: "<room shape/type>"}\\
\hspace*{1.0em}\texttt{grid\_alignment: "<what +Y/+X is
aligned with>"}\\
\hspace*{1.0em}\texttt{initial\_camera\_heading: "<heading
relative to room grid>"}\\[0.4em]

\textbf{Trajectory (example).}\\
\texttt{Trajectory:}\\
\hspace*{1.0em}\texttt{\# Track movement relative to the ROOM
GRID, not just camera view.}\\
\hspace*{1.0em}\texttt{- step: 0}\\
\hspace*{2.0em}\texttt{time: "0s"}\\
\hspace*{2.0em}\texttt{pos: [0.0, 0.0]}\\
\hspace*{2.0em}\texttt{facing: "NW (-X,+Y)"}\\
\hspace*{2.0em}\texttt{action: "Standing near entrance, panning across the room"}\\[0.2em]
\hspace*{1.0em}\texttt{- step: 1}\\
\hspace*{2.0em}\texttt{time: "4s"}\\
\hspace*{2.0em}\texttt{pos: [0.0, 1.8]}\\
\hspace*{2.0em}\texttt{facing: "North (+Y)"}\\
\hspace*{2.0em}\texttt{action: "s forward along the main room axis"}\\[0.2em]
\hspace*{1.0em}\texttt{- step: 2}\\
\hspace*{2.0em}\texttt{time: "8s"}\\
\hspace*{2.0em}\texttt{pos: [0.2, 3.5]}\\
\hspace*{2.0em}\texttt{facing: "East (+X)"}\\
\hspace*{2.0em}\texttt{action: "Turning right to inspect bedside area"}\\[0.4em]

\textbf{Entity\_Registry (example).}\\
\hspace*{1.0em}\texttt{\# The Map. Coordinates are strictly
[x, y] in meters.}\\
\hspace*{1.0em}\texttt{- id: "door\_01"}\\
\hspace*{2.0em}\texttt{category: "door"}\\
\hspace*{2.0em}\texttt{first\_seen\_at: "0s"}\\
\hspace*{2.0em}\texttt{state: "open"}\\
\hspace*{2.0em}\texttt{estimated\_pos: [0.8, 0.0]}\\
\hspace*{2.0em}\texttt{approx\_size: [0.9, 2.1, 0.1]}\\
\hspace*{2.0em}\texttt{visual\_signature: "White hinged door with silver handle"}\\
\hspace*{2.0em}\texttt{spatial\_relation: "At the entrance boundary of the bedroom"}\\[0.2em]
\hspace*{1.0em}\texttt{- id: "bed\_01"}\\
\hspace*{2.0em}\texttt{category: "bed"}\\
\hspace*{2.0em}\texttt{first\_seen\_at: "5s"}\\
\hspace*{2.0em}\texttt{estimated\_pos: [1.8, 2.8]}\\
\hspace*{2.0em}\texttt{approx\_size: [1.6, 2.0, 0.6]}\\
\hspace*{2.0em}\texttt{orientation: "Headboard against +X wall"}\\
\hspace*{2.0em}\texttt{visual\_signature: "Double bed with white sheets and dark frame"}\\
\hspace*{2.0em}\texttt{spatial\_relation: "Against the right wall, beside nightstand\_01"}\\[0.2em]
\hspace*{1.0em}\texttt{- id: "nightstand\_01"}\\
\hspace*{2.0em}\texttt{category: "nightstand"}\\
\hspace*{2.0em}\texttt{first\_seen\_at: "7s"}\\
\hspace*{2.0em}\texttt{estimated\_pos: [1.9, 2.0]}\\
\hspace*{2.0em}\texttt{approx\_size: [0.5, 0.6, 0.4]}\\
\hspace*{2.0em}\texttt{visual\_signature: "Small wooden bedside table"}\\
\hspace*{2.0em}\texttt{spatial\_relation: "In front of bed\_01 near the headboard"}\\[0.2em]
\hspace*{1.0em}\texttt{- id: "trash\_bin\_01"}\\
\hspace*{2.0em}\texttt{category: "trash\_bin"}\\
\hspace*{2.0em}\texttt{first\_seen\_at: "10s"}\\
\hspace*{2.0em}\texttt{estimated\_pos: [-1.3, 2.4]}\\
\hspace*{2.0em}\texttt{approx\_size: [0.3, 0.4, 0.3]}\\
\hspace*{2.0em}\texttt{visual\_signature: "Black cylindrical trash bin"}\\
\hspace*{2.0em}\texttt{spatial\_relation: "Near the left wall below desk\_01"}
\end{quote}

\begin{promptbox}{Textual Representation of Allocentric Context Prompt}
\label{trace_prompt}
\textbf{[System Prompt]}\\
You are an expert specialized in solving spatial understanding questions using text descriptions of egocentric video sequences.

You are given:\\
\noindent\hspace*{0.5em}-- an egocentric video of an indoor environment,\\
\noindent\hspace*{0.5em}-- a question about that video.

You need to firstly generate a structured ``Textual Representation of Allocentric Context'' derived from the egocentric video, and then answer the question as accurately as possible.\\
\smallskip

HOW TO GENERATE THE TEXTUAL REPRESENTATION OF ALLOCENTRIC CONTEXT:\\

1. Coordinate System Rules (Room-Aligned Allocentric Frame)\\
\noindent\hspace*{0.5em}-- Origin: the camera starting position is exactly \texttt{[0.0, 0.0]} on the floor plane.\\
\noindent\hspace*{0.5em}-- Major Axes (+Y / +X): Align the coordinate system with the \emph{dominant walls} or \emph{floor grid} of the room rather than the camera’s initial viewing direction.:\\
\noindent\hspace*{1.5em}-- define \texttt{`+Y'} along that dominant structural direction;\\
\noindent\hspace*{1.5em}-- define \texttt{`+X'} as the perpendicular rightward direction in the floor plane.\\
\noindent\hspace*{0.5em}-- Units are approximate meters.\\
\noindent\hspace*{0.5em}-- Maintain one globally consistent coordinate frame throughout the video. If the camera moves into another room, preserve the same global frame rather than resetting coordinates.\\
\smallskip

2. Meta-Context Rules\\
You must infer and report:\\
\noindent\hspace*{0.5em}-- \texttt{room\_topology}: the overall spatial structure of the observed environment, such as `rectangular bedroom`, `L-shaped office`, or `narrow hallway connected to kitchen`\\
\noindent\hspace*{0.5em}-- \texttt{grid\_alignment}: the structural cue used to define the allocentric axes\\
\noindent\hspace*{0.5em}-- \texttt{initial\_camera\_heading}: the camera’s initial facing direction relative to the room-aligned grid\\
\smallskip

3. Trajectory Rules\\
You must log the camera path continuously. Output a trajectory step for \emph{every significant} camera movement.\\
\noindent\hspace*{0.5em}-- \texttt{step}: Sequential ID.\\
\noindent\hspace*{0.5em}-- \texttt{time}: Timestamp of the step (e.g., "2s")
\noindent\hspace*{0.5em}-- \texttt{pos}: Estimated [x, y] of the camera.\\
\noindent\hspace*{0.5em}-- \texttt{facing}: Cardinal direction and axis (e.g., "North (+Y)").\\
\noindent\hspace*{0.5em}-- \texttt{action}: Short description of the camera motion or viewpoint change\\
\smallskip

4. Entity Registry Rules\\
You must register every visible entity individually. Never group objects. For each entity, include:\\
\noindent\hspace*{0.5em}-- \texttt{id}: unique identifier such as \texttt{chair\_01}, \texttt{door\_01}\\
\noindent\hspace*{0.5em}-- \texttt{category}\\
\noindent\hspace*{0.5em}-- \texttt{first\_seen\_at}\\
\noindent\hspace*{0.5em}-- \texttt{estimated\_pos}: \texttt{[x, y]}\\
\noindent\hspace*{0.5em}-- \texttt{approx\_size}: \texttt{[width, height, depth]}\\
\noindent\hspace*{0.5em}-- \texttt{visual\_signature}: short appearance-based description for disambiguation\\
\noindent\hspace*{0.5em}-- \texttt{spatial\_relation}: at least one relation to a nearby anchor or structure\\
\noindent\hspace*{0.5em}-- optional \texttt{state} when applicable (e.g., door open/closed, drawer open/closed)\\
\noindent\hspace*{0.5em}-- optional \texttt{orientation} when meaningful (e.g., bed headboard against east wall)\\
\smallskip

Output Format (Strict YAML)\\
\emph{refer to above}\\
\smallskip

Hard Rules\\
\noindent\hspace*{0.5em}-- No omissions: if an object occupies more than 1\% of pixels, it must be listed (including partial window/door edges).\\
\noindent\hspace*{0.5em}-- Force coordinates: you must estimate \texttt{[x,y]} for every listed item.\\
\noindent\hspace*{0.5em}-- Exhaustive count: never group items; if there are 6 chairs, I expect 6 entries in the registry.\\
\noindent\hspace*{0.5em}-- Global consistency: trajectory and entity coordinates must agree with one another.
\smallskip

Answer format:\\
- \textsc{\{POST\_PROMPT\}}\\
- Do NOT add extra text on the final answer line (no units, no explanations).\\
\smallskip

\textbf{[User Prompt]}\\
\textbf{Reasoning protocol:}\\
\noindent\hspace*{0.5em}-- Read the question and analyze the visual content carefully.\\
\noindent\hspace*{0.5em}-- Generate the textual representation of allocentric context as required to determine spatial relations and approximate distances.\\
\noindent\hspace*{0.5em}-- You may think through the problem internally, but do \emph{not} show your reasoning.\\
\noindent\hspace*{0.5em}-- Directly provide the final answer in the required format.\\
\smallskip

\textbf{Output format:}\\
\noindent Structured Textual Allocentric Representation:\\
\noindent \\
\noindent Answer: \texttt{`X'}\\
\smallskip

Now follow this protocol to answer the question below.
\end{promptbox}

\section{Additional Results}
\label{app:evaluation}

\subsection{Details on Decomposition Analysis}

\begin{table*}[t]
\centering
\scriptsize
\captionsetup{skip=6pt}
\caption{\textit{Decomposition analysis of 3D spatial QA performance (matrix).} We adopt the text-only prediction setting and report results for different combinations of visual descriptors and spatial knowledge parsers. Cell shows Avg.\ with (Multiple-Choice Answer / Numerical Answer).}
\label{tab:decompose_small}
\setlength{\tabcolsep}{2pt}
\renewcommand{\arraystretch}{1.10}
\begin{tabularx}{0.98\linewidth}{@{}
>{\raggedright\arraybackslash}p{3cm}
*{4}{>{\centering\arraybackslash}X}
@{}}
\toprule
Descriptor $\backslash$ Parser
& \shortstack{Gemini 3 Pro\\Avg.\ (MCA/NA)}
& \shortstack{Qwen2.5-72B-Instruct\\Avg.\ (MCA/NA)}
& \shortstack{Qwen2.5-32B-Instruct\\Avg.\ (MCA/NA)}
& \shortstack{Qwen2.5-7B-Instruct\\Avg.\ (MCA/NA)} \\
\midrule
Gemini 3 Pro
& \shortstack{52.27\\(44.12/57.40)}
& \shortstack{40.86\\(42.33/39.47)}
& \shortstack{36.23\\(38.03/41.35)}
& \shortstack{29.35\\(30.00/28.73)} \\
Qwen2.5-VL-72B-Instruct
& \shortstack{36.11\\(41.58/28.94)}
& \shortstack{31.11\\(35.98/26.51)}
& \shortstack{26.12\\(32.01/20.56)}
& \shortstack{24.06\\(29.60/18.84)} \\
Qwen2.5-VL-32B-Instruct
& \shortstack{36.70\\(40.74/32.18)}
& \shortstack{14.66\\(29.20/0.94)}
& \shortstack{23.29\\(29.60/17.34)}
& \shortstack{24.45\\(30.24/18.99)} \\
Qwen2.5-VL-7B-Instruct
& \shortstack{32.72\\(36.38/28.39)}
& \shortstack{29.08\\(32.85/25.53)}
& \shortstack{24.33\\(29.00/19.92)}
& \shortstack{25.19\\(29.28/21.34)} \\
\bottomrule
\end{tabularx}
\end{table*}

\begin{table*}[t]
\centering
\scriptsize
\captionsetup{skip=6pt}
\caption{\textit{Decomposition analysis of 3D spatial QA performance (grouped breakdown).}
We adopt the text-only prediction setting and report results for different combinations of visual descriptors and spatial knowledge parsers.}
\label{tab:decompose_big}
\setlength{\tabcolsep}{2pt}
\renewcommand{\arraystretch}{1.10}
\begin{tabularx}{0.98\linewidth}{@{}
>{\raggedright\arraybackslash}p{2cm}
>{\raggedright\arraybackslash}p{1.9cm}
>{\centering\arraybackslash}p{1cm}
 *{4}{>{\centering\arraybackslash}X}
 *{4}{>{\centering\arraybackslash}X}
@{}}
\toprule
\multirow{2}{*}{Descriptor} &
\multirow{2}{*}{Parser} &
\multirow{2}{*}{Avg.} &
\multicolumn{4}{c}{Numerical Answer} &
\multicolumn{4}{c}{Multiple-Choice Answer} \\
\cmidrule(lr){4-7}\cmidrule(lr){8-11}
& & &
Obj. Cnt. & Abs. Dist. & Obj. Size & Room Size &
Rel. Dist. & Rel. Dir. & Route & Order \\
\midrule

\multirow{4}{*}{Gemini 3 Pro}
& Gemini 3 Pro   & 52.27 & 28.52 & 32.66 & 67.28 & 48.02 & 49.58 & 62.66 & 52.43 & 64.93 \\
& Qwen2.5-72B    & 40.86 & 28.12 & 28.91 & 55.38 & 39.69 & 39.30 & 40.60 & 34.54 & 50.97 \\
& Qwen2.5-32B    & 36.23 & 27.45 & 21.55 & 49.81 & 44.10 & 40.00 & 30.99 & 30.41 & 45.15 \\
& Qwen2.5-7B     & 29.35 & 24.62 & 22.95 & 36.13 & 29.03 & 28.87 & 28.31 & 26.29 & 35.11 \\
\midrule

\multirow{4}{*}{Qwen2.5-VL-72B}
& Gemini 3 Pro   & 36.11 & 10.57 & 16.57 & 51.83 & 31.72 & 40.25 & 39.93 & 34.81 & 47.39 \\
& Qwen2.5-72B    & 31.11 & 12.83 & 21.74 & 39.71 & 23.51 & 37.04 & 33.16 & 32.99 & 40.13 \\
& Qwen2.5-32B    & 26.12 & 11.86 & 13.49 & 32.03 & 20.17 & 35.07 & 27.17 & 32.47 & 35.92 \\
& Qwen2.5-7B     & 24.06 & 11.10 & 16.26 & 25.54 & 19.34 & 29.01 & 27.58 & 27.32 & 34.14 \\
\midrule

\multirow{4}{*}{Qwen2.5-VL-32B}
& Gemini 3 Pro   & 36.70 & 14.93 & 19.07 & 52.20 & 38.20 & 36.93 & 39.58 & 34.27 & 48.78 \\
& Qwen2.5-72B    & 14.66 & 0.00  & 2.87  & 0.10  & 0.00  & 35.21 & 30.68 & 33.51 & 19.42 \\
& Qwen2.5-32B    & 23.29 & 0.00  & 13.45 & 36.26 & 0.00  & 29.86 & 40.70 & 31.96 & 11.49 \\
& Qwen2.5-7B     & 24.45 & 0.00  & 27.75 & 20.45 & 26.08 & 29.15 & 38.74 & 33.51 & 17.15 \\
\midrule

\multirow{4}{*}{Qwen2.5-VL-7B}
& Gemini 3 Pro   & 32.72 & 10.81 & 14.11 & 49.46 & 34.54 & 37.77 & 37.83 & 31.46 & 33.99 \\
& Qwen2.5-72B    & 29.08 & 11.96 & 20.92 & 36.59 & 28.85 & 30.00 & 35.74 & 32.47 & 31.72 \\
& Qwen2.5-32B    & 24.33 & 11.43 & 13.72 & 29.62 & 22.43 & 30.28 & 28.51 & 32.99 & 27.02 \\
& Qwen2.5-7B     & 25.19 & 12.02 & 21.63 & 27.51 & 18.40 & 29.44 & 29.86 & 30.41 & 27.83 \\
\bottomrule
\end{tabularx}
\end{table*}

We present the per-category and per-task breakdowns of our compositional analysis in Tables~\ref{tab:decompose_small} and~\ref{tab:decompose_big}, respectively. These results further reveal how different descriptor–parser combinations contribute to spatial reasoning performance and highlight key bottlenecks in modeling object-, relation-, and layout-level spatial concepts. 

\subsection{Instruction Following for Spatial Reasoning}

\begin{table*}[t]
\centering
\scriptsize
\captionsetup{skip=6pt}
\caption{Effect of visual tokens and multimodal training on spatial context reasoning.}
\label{tab:analysis}
\setlength{\tabcolsep}{2pt}
\renewcommand{\arraystretch}{1.10}
\begin{tabularx}{0.98\linewidth}{@{}
>{\raggedright\arraybackslash}p{2cm}
>{\raggedright\arraybackslash}p{2cm}
>{\centering\arraybackslash}p{0.9cm}
*{4}{>{\centering\arraybackslash}X}
*{4}{>{\centering\arraybackslash}X}
@{}}
\toprule
\multirow{2}{*}{Descriptor} &
\multirow{2}{*}{Parser} &
\multirow{2}{*}{Avg.} &
\multicolumn{4}{c}{Numerical Answer} &
\multicolumn{4}{c}{Multiple-Choice Answer} \\
\cmidrule(lr){4-7} \cmidrule(lr){8-11}
& & &
Obj. Cnt. & Abs. Dist. & Obj. Size & Room Size &
Rel. Dist. & Rel. Dir. & Route & Order \\
\midrule

Gemini 3 Pro     & Qwen2.5-72B       & 40.86 & 28.12 & 28.91 & 55.38 & 39.69 & 39.30 & 40.60 & 34.54 & 50.97 \\
Gemini 3 Pro     & Qwen2.5-VL-72B    & 39.48 & 28.21 & 20.56 & 54.23 & 42.15 & 44.37 & 34.50 & 39.18 & 53.56 \\
Qwen2.5-VL-72B   & Qwen2.5-72B       & 31.11 & 12.83 & 21.74 & 39.71 & 23.51 & 37.04 & 33.16 & 32.99 & 40.13 \\
Qwen2.5-VL-72B   & Qwen2.5-VL-72B    & 26.65 & 9.08  & 11.08 & 36.21 & 20.59 & 28.31 & 30.58 & 36.08 & 40.78 \\
\bottomrule
\end{tabularx}
\end{table*}

In Tab.~\ref{tab:analysis}, we conduct additional experiments comparing Qwen-VL and Qwen LLM under the same textual representations. We find that Qwen-VL consistently underperforms the language-only model for both Gemini-generated and Qwen-generated representations. This observation suggests that visual instruction tuning can compromise spatial knowledge parsing ability, highlighting the importance of carefully designing visual training data to enable MLLMs to better capture 3D spatial concepts.

\subsection{Detailed Results on Stratified Analysis}

\begin{table*}[t]
\centering
\footnotesize
\captionsetup{skip=6pt}
\caption{\textit{Stratified VSI-Bench results across underlying scene datasets.} Performance is broken down by ARKitScenes, ScanNet, and ScanNetPP for both proprietary and open-weight models. TRACE consistently improves over the Direct baseline across all environment distributions, indicating strong cross-environment generalization.}
\label{tab:stratified}
\setlength{\tabcolsep}{2pt}
\renewcommand{\arraystretch}{1.10}
\begin{tabularx}{0.98\linewidth}{@{}
>{\raggedright\arraybackslash}p{2.5cm}
>{\centering\arraybackslash}p{2cm}
*{3}{>{\centering\arraybackslash}X}
@{}}
\toprule
Method & Avg. & ARKitScenes & ScanNet & ScanNetPP \\
\midrule

\multicolumn{5}{@{}l@{}}{\textit{o3 as base model}} \\
Direct & 51.15 & 49.28 & 53.55 & 49.78 \\
CoT    & 52.36 & 51.76 & 53.53 & 51.36 \\
ToT    & 52.09 & 51.27 & 53.46 & 51.06 \\
LtM    & 52.50 & 51.70 & 54.31 & 50.80 \\
CM     & 53.93 & 52.69 & \textbf{56.24} & 52.00 \\
Ours   & \textbf{54.08} & \textbf{54.63} & 55.06 & \textbf{52.11} \\
\midrule

\multicolumn{5}{@{}l@{}}{\textit{MiMo-VL-7B-SFT as base model}} \\
Direct & 39.79 & 39.36 & 39.97 & \textbf{40.01} \\
CoT    & 37.49 & 37.33 & 37.08 & 38.26 \\
ToT    & 39.14 & 38.45 & 40.50 & 37.97 \\
LtM    & 38.34 & 36.52 & 39.52 & 38.66 \\
CM     & 36.85 & 36.31 & 36.99 & 37.23 \\
Ours   & \textbf{41.42} & \textbf{42.47} & \textbf{41.72} & 39.82 \\
\midrule

\multicolumn{5}{@{}l@{}}{\textit{Qwen2.5-VL-72B-Instruct as base model}} \\
Direct & 36.28 & 36.05 & 36.20 & 36.65 \\
CoT    & 29.78 & 27.76 & 31.37 & 29.74 \\
ToT    & 38.06 & 40.43 & 36.84 & 37.19 \\
LtM    & 38.01 & 40.41 & 36.74 & 37.18 \\
CM     & 35.47 & 37.83 & 35.07 & 33.45 \\
Ours   & \textbf{39.38} & \textbf{42.03} & \textbf{37.80} & \textbf{38.73} \\
\midrule

\multicolumn{5}{@{}l@{}}{\textit{GLM-4.5V as base model}} \\
Direct & 37.33 & 33.97 & 38.58 & 39.26 \\
CoT    & 38.48 & 36.58 & 39.81 & 38.73 \\
ToT    & 40.66 & 39.07 & 41.44 & 41.30 \\
LtM    & 40.99 & 39.16 & 42.13 & 41.44 \\
CM     & 38.93 & 37.92 & 39.03 & 39.89 \\
Ours   & \textbf{45.01} & \textbf{46.83} & \textbf{44.37} & \textbf{43.89} \\
\midrule

\multicolumn{5}{@{}l@{}}{\textit{Gemini 3 Pro as base model}} \\
Direct & 52.61 & 52.26 & 52.61 & 52.99 \\
CoT    & 53.65 & 52.45 & 53.68 & 54.93 \\
ToT    & 58.88 & 57.61 & 58.32 & \textbf{61.10} \\
LtM    & 59.52 & 58.01 & 59.78 & 60.83 \\
CM     & 59.72 & \textbf{59.50} & 59.54 & 60.23 \\
Ours   & \textbf{60.15} & 59.39 & \textbf{60.42} & 60.63 \\
\bottomrule
\end{tabularx}
\end{table*}

Tab.~\ref{tab:stratified} provides a granular breakdown of model performance across the three distinct indoor scene datasets comprising VSI-Bench: ARKitScenes~\cite{baruch2021arkitscenes}, ScanNet~\cite{dai2017scannet}, and ScanNetPP~\cite{yeshwanth2023scannet++}. Across both proprietary and open-weights architectures, our proposed TRACE prompting robustly yields performance improvements over the Direct baseline within each environment distribution. Notably, TRACE achieves balanced gains without overfitting to a specific dataset's spatial characteristics, confirming the reliable cross-environment generalization of our textual allocentric representation.

\subsection{Full Evaluation Results on VSI-Bench}

\definecolor{oursgreen}{RGB}{220,245,220}

\begin{table*}[t]
\captionsetup{skip=6pt}
\caption{\textit{Evaluation results on the VSI benchmark.} We report average performance and detailed breakdowns across numerical-answer and multiple-choice tasks, under proprietary and open-sourced base models. Best results are in \textbf{bold}, and second-best are \underline{underlined}.}
\label{tab:vsi_full}
\centering
\setlength{\tabcolsep}{3.5pt}
\scriptsize
\renewcommand{\arraystretch}{1.0}
\begin{tabularx}{\linewidth}{@{} L{1.2} C{1.1} *{8}{C{0.96}} @{}}
\toprule
\multirow{2}{*}{Methods} &
\multirow{2}{*}{Avg.} &
\multicolumn{4}{c}{\cellcolor{yellow!35}Numerical Answer} &
\multicolumn{4}{c@{}}{\cellcolor{cyan!25}Multiple-Choice Answer} \\
\cmidrule(r){3-6} \cmidrule(l){7-10}
& &
\cellcolor{yellow!12}\makebox[0pt][c]{Obj.\ Cnt.} &
\cellcolor{yellow!12}\makebox[0pt][c]{Abs.\ Dist.} &
\cellcolor{yellow!12}\makebox[0pt][c]{Obj.\ Size} &
\cellcolor{yellow!12}\makebox[0pt][c]{Room Size} &
\cellcolor{cyan!10}\makebox[0pt][c]{Rel.\ Dist.} &
\cellcolor{cyan!10}\makebox[0pt][c]{Rel.\ Dir.} &
\cellcolor{cyan!10}\makebox[0pt][c]{Route} &
\cellcolor{cyan!10}\makebox[0pt][c]{Order} \\
\midrule
\multicolumn{10}{@{}l@{}}{\textit{Gemini 3 Pro as base model}} \\
Direct & 52.61 & 33.77 & 32.57 & 67.09 & 42.99 & 62.54 & 50.52 & 51.03 & 70.71 \\
CoT    & 53.65 & 30.35 & 34.54 & 64.05 & 40.76 & 61.78 & 58.09 & \textbf{61.34} & 71.96 \\
ToT    & 58.88 & 44.55 & \textbf{42.12} & 72.20 & 45.55 & \underline{65.35} & 57.83 & 55.62 & \textbf{73.73} \\
LtM    & 59.52 & 45.19 & 40.72 & \underline{73.36} & 44.15 & \textbf{65.82} & \underline{60.40} & 53.59 & \underline{73.64} \\
CM     & \underline{59.72} & \underline{46.70} & \underline{41.43} & 72.49 & \textbf{50.14} & 63.69 & 58.62 & 55.50 & 72.61 \\
\rowcolor{oursgreen}
Ours   & \textbf{60.15} & \textbf{47.55} & 38.82 & \textbf{73.90} & \underline{45.62} & 63.85 & \textbf{61.70} & \underline{58.01} & 72.97 \\
\midrule
\multicolumn{10}{@{}l@{}}{\textit{Qwen2.5-VL-72B-Instruct as base model}} \\
Direct & 36.28 & \textbf{33.36} & 20.53 & 49.31 & \underline{41.49} & \textbf{43.38} & 27.79 & 32.47 & \textbf{44.01} \\
CoT    & 29.78 & 21.27 & 24.95 & 16.31 & 40.94 & 39.44 & 33.16 & 28.87 & \underline{43.53} \\
ToT    & \underline{38.06} & 17.89 & 26.20 & 53.15 & \textbf{47.01} & 41.55 & \underline{36.78} & \textbf{35.05} & \textbf{44.01} \\
LtM    & 38.01 & \underline{23.27} & \textbf{31.39} & \underline{54.49} & 38.68 & \underline{42.96} & 34.71 & 29.90 & 36.73 \\
CM     & 35.47 & 21.58 & 15.67 & 52.65 & 37.26 & 39.44 & 36.05 & \underline{34.54} & 42.39 \\
\rowcolor{oursgreen}
Ours   & \textbf{39.38} & 22.05 & \underline{28.03} & \textbf{59.98} & 38.99 & 40.85 & \textbf{37.40} & 31.96 & 42.56 \\
\midrule
\multicolumn{10}{@{}l@{}}{\textit{MiMo-VL-7B as base model}} \\
Direct & \underline{39.79} & \textbf{36.02} & 29.84 & 52.38 & 42.95 & 40.14 & \underline{33.78} & 31.44 & 47.41 \\
CoT    & 37.49 & 34.27 & 23.50 & 48.52 & \underline{43.23} & 38.73 & 32.75 & 27.84 & 49.23 \\
ToT    & 39.14 & 29.45 & \underline{30.44} & \underline{54.26} & 40.14 & \underline{41.41} & 32.02 & \underline{32.47} & 46.60 \\
LtM    & 38.34 & \underline{35.09} & 24.47 & 48.22 & \textbf{44.48} & \textbf{43.10} & 30.79 & \textbf{35.05} & \underline{49.50} \\
CM     & 36.85 & 27.43 & 23.14 & 50.14 & 39.06 & \underline{41.41} & 32.54 & 27.84 & 46.76 \\
\rowcolor{oursgreen}
Ours   & \textbf{41.42} & 33.27 & \textbf{31.51} & \textbf{58.67} & 41.56 & 39.44 & \textbf{35.33} & 28.87 & \textbf{51.29} \\
\midrule
\multicolumn{10}{@{}l@{}}{\textit{o3 as base model}} \\
Direct & 51.15 & 33.26 & \underline{31.95} & 69.37 & \underline{52.57} & 58.87 & 44.11 & 42.78 & 69.42 \\
CoT    & 52.36 & 34.11 & 28.37 & 69.81 & 50.31 & \textbf{59.72} & 48.89 & 57.06 & 70.96 \\
ToT    & 52.09 & \underline{40.07} & 24.26 & 69.55 & 48.68 & 59.15 & 50.23 & 55.35 & 69.36 \\
LtM    & 52.50 & 35.68 & 26.98 & 70.05 & 47.05 & 59.15 & \underline{50.97} & \underline{57.96} & \underline{71.22} \\
CM     & \underline{53.93} & 34.18 & \textbf{33.35} & \underline{70.19} & 52.05 & \underline{59.30} & \textbf{51.10} & \textbf{62.99} & \textbf{71.26} \\
\rowcolor{oursgreen}
Ours   & \textbf{54.08} & \textbf{43.40} & 29.93 & \textbf{72.48} & \textbf{54.03} & 57.32 & 49.83 & 56.10 & 70.02 \\
\midrule
\multicolumn{10}{@{}l@{}}{\textit{GLM-4.5V as base model}} \\
Direct & 37.33 & 34.87 & 32.74 & 28.13 & 29.72 & \underline{47.32} & 39.05 & 35.57 & 49.92 \\
CoT    & 38.48 & 33.42 & 31.07 & 39.88 & 25.52 & 45.49 & \underline{39.26} & 35.57 & 50.19 \\
ToT    & 40.66 & 34.45 & 32.17 & 47.29 & 27.81 & 45.63 & \textbf{39.77} & 32.47 & 51.88 \\
LtM    & \underline{41.35} & 32.32 & \textbf{33.14} & \underline{51.26} & 22.26 & \textbf{49.30} & 36.47 & \textbf{39.18} & \textbf{58.00} \\
CM     & 38.93 & \underline{37.77} & 31.91 & 36.61 & \underline{30.07} & 45.21 & 37.40 & 33.51 & \underline{54.05} \\
\rowcolor{oursgreen}
Ours   & \textbf{45.01} & \textbf{40.41} & \underline{32.84} & \textbf{65.11} & \textbf{36.74} & 45.77 & 38.53 & \underline{36.60} & 50.68 \\
\bottomrule
\end{tabularx}
\end{table*}

The complete results for VSI-Bench are detailed in Table~\ref{tab:vsi_full}. To mitigate the risk of data contamination, we restrict our evaluation to model versions released no later than six months after the publication of VSI-Bench and OST-Bench. Consequently, our final selection includes Gemini~3~Pro~\cite{geminiteam2025geminifamilyhighlycapable}, o3~\cite{openai2025o3}, Qwen2.5-VL~\cite{qwen2.5}, MiMo-VL-7B~\cite{coreteam2025mimovltechnicalreport}, and GLM-4.5V~\cite{vteam2025glm45vglm41vthinkingversatilemultimodal}.

\subsection{Token Efficiency}

\begin{table*}[t]
\centering
\small
\caption{\textit{Token consumption and performance trade-offs of prompting methods across different models.} TRACE generally maintains a favorable performance-to-cost ratio relative to branching reasoning baselines, although its token usage varies by backbone. Best results for each model aredo in \textbf{bold}.}
\label{tab:token}
\setlength{\tabcolsep}{3pt}
\renewcommand{\arraystretch}{1.10}
\sisetup{reset-text-series=false, text-series-to-math=true}
\begin{tabularx}{0.98\linewidth}{@{}l *{5}{>{\centering\arraybackslash}X >{\centering\arraybackslash}X}@{}}
\toprule
& \multicolumn{2}{c}{GLM-4.5V}
& \multicolumn{2}{c}{MiMo-VL-7B}
& \multicolumn{2}{c}{Qwen2.5-VL-72B}
& \multicolumn{2}{c}{o3}
& \multicolumn{2}{c}{Gemini 3 Pro} \\
\cmidrule(lr){2-3}\cmidrule(lr){4-5}\cmidrule(lr){6-7}\cmidrule(lr){8-9}\cmidrule(lr){10-11}
Method & Tok & Avg
& Tok & Avg
& Tok & Avg
& Tok & Avg
& Tok & Avg \\
\midrule
Direct & 405.17 & 37.33 & 337.36 & 39.79 & 3.48   & 36.28 & 3.61   & 51.15 & 334.35 & 52.61 \\
CoT    & 568.36 & 38.48 & 579.21 & 37.49 & 129.62 & 28.44 & 76.20  & 52.36 & 479.64 & 53.65 \\
ToT    & 1079.97 & 40.66 & 1132.86 & 39.14 & 308.99 & 37.68 & 352.30 & 52.09 & 450.82 & 58.88 \\
LtM    & 989.55 & 40.99 & 1097.05 & 38.34 & 229.84 & 38.01 & 220.28 & 52.50 & 571.88 & 59.52 \\
CM     & 722.16 & 38.93 & 723.68 & 36.85 & 224.72 & 35.47 & 81.07  & 53.93 & 403.23 & 59.72 \\
\rowcolor{gray!8}
\textbf{Ours} & 967.91 & \bfseries 45.01 & 737.72 & \bfseries 41.42 & 755.87 & \bfseries 39.38 & 435.49 & \bfseries 54.08 & 843.91 & \bfseries 60.15 \\
\bottomrule
\end{tabularx}
\label{tab:main_results}
\end{table*}

As shown in the Tab.~\ref{tab:token}, the token consumption of TRACE varies depending on the underlying model's generation tendencies but generally maintains a highly favorable performance-to-cost trade-off, particularly when compared to highly branching reasoning methods. For instance, on the compact MiMo-VL-7B model, TRACE consumes significantly fewer tokens (737.72) than Tree-of-Thoughts (1132.86) and Least-to-Most (1097.05) prompting, while simultaneously delivering superior average performance. However, for several other large foundation models, including Gemini 3 Pro, o3, and Qwen2.5-VL-72B, our method is noticeably more token-intensive than these baselines. This increased consumption is an expected trade-off, as explicitly generating a structured allocentric representation inherently loads the context window with an exhaustive spatial cache. While the consistent accuracy gains across diverse models justify this computational overhead, optimizing token efficiency during the structured reasoning process constitutes a largely orthogonal research direction, which we leave for future work.

\end{document}